\pdfoutput=1

\documentclass[11pt, table]{article}

\usepackage[preprint]{acl}

\usepackage{upquote}
\usepackage{enumitem}

\usepackage{xcolor}

\usepackage{multirow}
\usepackage{makecell}

\usepackage{times}
\usepackage{latexsym}
\usepackage{arabtex}
\usepackage{utf8}
\setcode{utf8}

\usepackage[most]{tcolorbox}
\usepackage{float}

\usepackage{booktabs}

\usepackage[T1]{fontenc}

\usepackage{microtype}

\usepackage{inconsolata}

\usepackage{graphicx}
\usepackage[font=small,labelfont=bf]{caption}

\definecolor{culturehighlight}{RGB}{255, 240, 220}
\definecolor{figurativehighlight}{RGB}{230, 240, 255}
\definecolor{poetryhighlight}{RGB}{230, 240, 255}

\makeatletter
\renewenvironment{abstract}%
  {\centerline{\large\bf Abstract}%
    \begin{list}{}%
      {\setlength{\rightmargin}{0.6cm}%
       \setlength{\leftmargin}{0.6cm}}%
     \item[]\ignorespaces%
     \@setsize\normalsize{12pt}\xpt\@xpt
  }%
  {\unskip\end{list}}
\makeatother

\title{Figurative and Cultural Knowledge in LLMs: Investigating Cross-Domain Transfer through Fine-Tuning}

\author{
 \textbf{Mena Attia\textsuperscript{1}},
 \textbf{Mona Diab\textsuperscript{2}},
 \textbf{Thamar Solorio\textsuperscript{1}}
\\
\\
 \textsuperscript{1}MBZUAI,
 \textsuperscript{2}Carnegie Mellon University
\\
 \texttt{\{mena.attia, thamar.solorio\}@mbzuai.ac.ae}
}

\begin{document}
\maketitle
\begin{abstract}

Figurative language is deeply culturally embedded; fluent use requires not just linguistic competence but cultural immersion. We ask whether LLMs can learn this link: does fine-tuning on cultural data improve figurative language understanding, and vice versa? We conduct a systematic study across four models (ALLaM-7B, Fanar-1-9B, Qwen3-8B, Llama-3.1-8B) and six Arabic datasets spanning cultural commonsense, proverbs, and poetry across diverse dialects and regions. Fine-tuning on poetry improves idiom comprehension ($+2.33\%$, $p<0.05$), a gain our ArabicMMLU control does not reproduce, indicating that it stems from figurative content rather than Arabic language adaptation and pointing to a sensitivity to non-literal meaning that transfers across figurative types. Cultural fine-tuning, by contrast, lowers proverb-interpretation accuracy in both Arabic-centric models. Transfer between the two domains is otherwise indistinguishable from noise, with Arabic models frequently regressing after fine-tuning, suggesting prior saturation of relevant knowledge, while multilingual models show greater adaptation headroom. Error analysis further reveals that fine-tuning reinforces experiential cultural knowledge while destabilizing historically grounded factual knowledge. Our findings suggest that the relationship between culture and figurative language, though conceptually natural, is not straightforwardly captured through fine-tuning alone.

\end{abstract}

\section{Introduction}

\begin{figure}[ht]
    \centering
    \includegraphics[width=0.95\columnwidth]{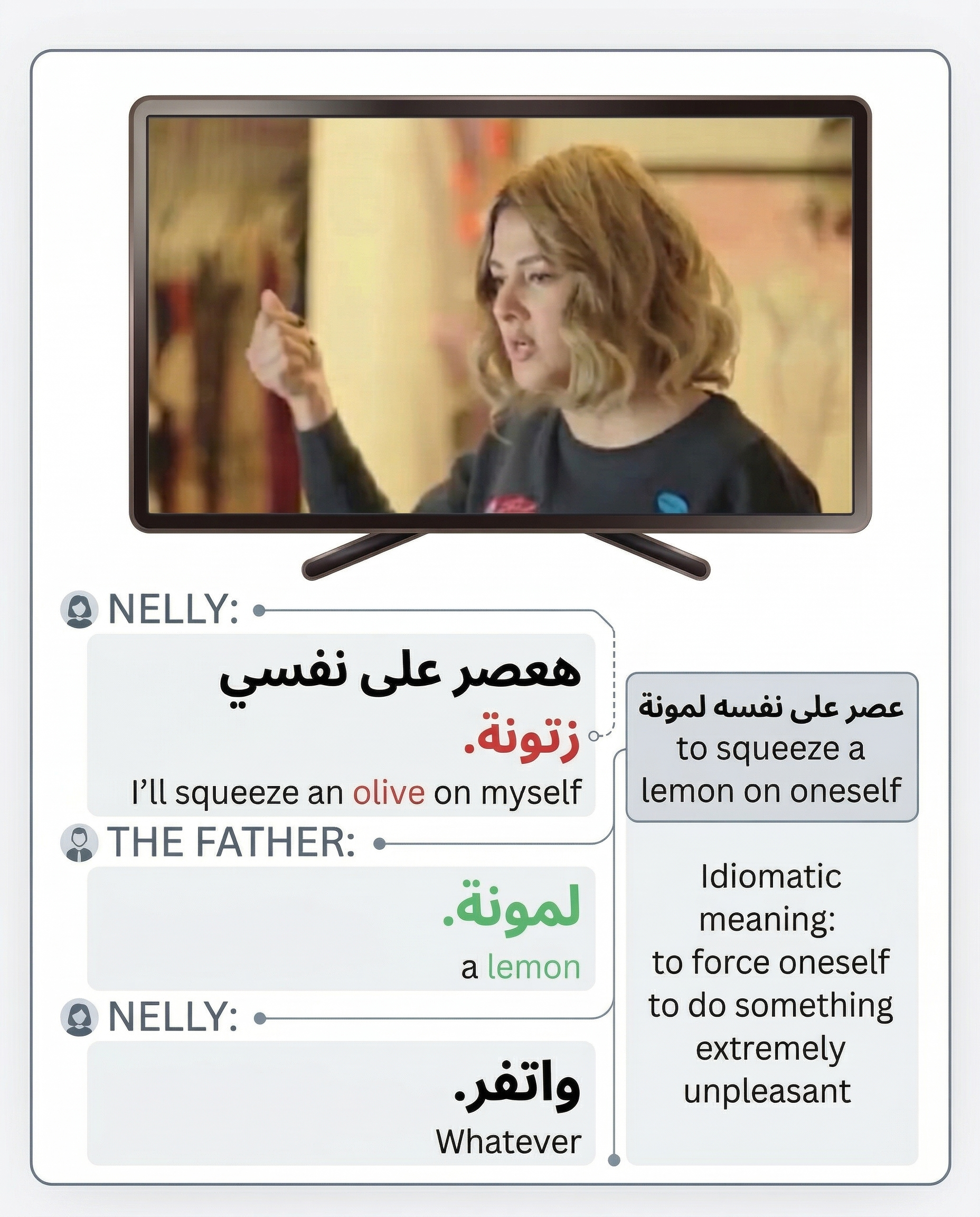}
    \caption{Scene from the Egyptian TV show Nelly and Sherihan.}
    \label{fig:nelly}
    \vspace{-0.2in}
\end{figure}

Figurative language is deeply embedded in culture. Proverbs and idioms, and poetry do more than convey non-literal meaning; they encode shared values, social norms, and historical experience, so that interpreting them draws on familiarity with the community that produced them rather than on linguistic competence alone. This premise is well established outside Natural Language Processing (NLP). Cognitive linguistics documents extensive cross-cultural variation in figurative meaning: even conceptual metaphors grounded in shared bodily experience are elaborated and conventionalized differently across speech communities \citep{Kovecses2005}, and cultural conceptualizations are encoded in and reproduced through its language \citep{sharifian2017cultural}. Conventional Figurative Language Theory likewise treats idioms as units whose interpretation draws on culture-specific connotations, so an idiom may be structurally transparent yet opaque to an outsider \citep{Dobrovolskij1756694}, and paremiology makes the same point for proverbs, which derive their force from a community's shared stock of experience rather than compositional meaning \citep{mieder2004proverbs,mieder1993proverbs}. Cognitive science converges on this: figurative interpretation recruits conceptual and background knowledge directly rather than extending literal decoding \citep{gibbs1994poetics}. Together, this literature motivates treating figurative competence and cultural knowledge as coupled rather than independent capacities.

The 2016 Egyptian television series \textit{Nelly and Sherihan} offers a concrete instantiation. Nelly, wealthy and westernized, repeatedly attempts Egyptian proverbs and misuses them, drawing corrections that she dismisses with ``whatever,'' while her working-class cousin wields the same expressions fluently (Figure~\ref{fig:nelly}). Its comedic device works only because the audience already shares the norm that misusing proverbs signals cultural distance, which is itself an attestation of the coupling.

In NLP, however, the two are studied separately. Cultural benchmarks evaluate factual knowledge, commonsense reasoning, social practices, and regional variation; figurative-language benchmarks evaluate the interpretation of proverbs, idioms, poetry, and metaphor. Whether the two forms of knowledge support one another in large language models remains untested: we do not know whether culturally grounded supervision improves figurative interpretation, or whether figurative supervision builds more general cultural knowledge.
 
We address this gap by investigating the relationship between cultural and figurative knowledge through controlled fine-tuning experiments in Arabic. We study two directions of cross-domain transfer. First, we ask whether fine-tuning on Arabic cultural data improves the interpretation of Arabic idioms and proverbs. Second, we ask whether fine-tuning on Arabic figurative-language data improves performance on culturally grounded question answering. We additionally examine transfer across figurative forms, asking whether supervision on proverbs or poetry improves understanding of unseen idioms and proverbs. Third, because both kinds of supervision are also exposure to Arabic, we ask whether any observed transfer reflects the content of the training data or Arabic-language adaptation alone. These experiments allow us to distinguish transfer between culture and figurative language from cross-form figurative transfer that may generalize across non-literal forms.

Our contributions are as follows. (1) We present a controlled framework for studying bidirectional transfer between Arabic cultural knowledge and figurative-language understanding. (2) We provide model-, dialect-, and topic-level analyses showing that fine-tuning effects are limited, unevenly distributed, and dependent on both model family and knowledge type. (3) We identify poetry fine-tuning as the only reliable source of positive figurative transfer, supporting the possibility of cross-form figurative transfer not restricted to a single non-literal category.
 
Taken together, our results show that the relationship between cultural and figurative knowledge is not straightforwardly captured through fine-tuning. Transfer effects are generally small, inconsistent, and highly model-dependent.

\section{Related Work}

\subsection{Benchmarks for Figurative Language and Culture}

\noindent\textbf{Global figurative benchmarks.} In English, Fig-QA
\cite{liu-etal-2022-testing} frames figurative interpretation as multiple-choice
question answering, while FLUTE \cite{chakrabarty-etal-2022-flute} provides
human-written explanations for idioms alongside other figurative categories.
Multilingual coverage is thinner. MABL \cite{kabra-etal-2023-multi} evaluates
figurative understanding across eight languages, MAPS
\cite{liu-etal-2024-multilingual} evaluates proverb understanding across multiple
languages, ProverbEval \cite{azime-etal-2025-proverbeval} covers cultural
proverbs in Ethiopian languages and English, and MIDI
\cite{almheiri-etal-2026-multilingual} spans idioms in 18 languages. The AdMIRe 2
shared task \cite{arslan-etal-2026-mwe} challenges models to interpret idiomatic
expressions across multiple languages.

\noindent\textbf{Arabic figurative benchmarks.} Three recent benchmarks target
Arabic: Jawaher \cite{magdy-etal-2025-jawaher}, a large-scale collection of
proverbs across many dialects; Kinayat \cite{attia-etal-2026-beyond}, a benchmark
of Egyptian Arabic idioms; and FannOrFlop
\cite{alghallabi2025fannflopmultigenremultiera}, a benchmark for poetry
interpretation spanning multiple genres and eras. We draw our training and
evaluation data from these resources (Section~\ref{sec:datasets}).

\noindent\textbf{Global cultural benchmarks.} BLEnD
\cite{myung2025blendbenchmarkllmseveryday} spans 16 regions and 13 languages with
52.6K QA pairs, and CulturalBench \cite{chiu-etal-2025-culturalbench} covers 45
regions through human--AI red-teaming, but both emphasize factual cultural
knowledge. ALM-Bench \cite{vayani2025languagesmatterevaluatinglmms} and CVQA
\cite{romero2024cvqaculturallydiversemultilingualvisual} extend coverage to 100
and 31 languages respectively, yet target visual and commonsense knowledge rather
than figurative interpretation.

\noindent\textbf{Arabic cultural benchmarks.} AraDiCE
\cite{mousi2024aradicebenchmarksdialectalcultural}, CAMEL-Bench
\cite{ghaboura2024camelbenchcomprehensivearabiclmm}, and CIDAR
\cite{alyafeai-etal-2024-cidar} cover dialectal variation, cultural knowledge, and
reasoning, but focus on factual or task-oriented understanding. Figurative
language appears only marginally elsewhere: ArabCulture
\cite{sadallah2025commonsensereasoningarabculture} covers 13 countries and 12
topics but allocates five idiom samples per country, and Palm
\cite{alwajih-etal-2025-palm} treats proverbs as one of 20 topics across 22
countries without in-depth evaluation. Across both Arabic and global benchmarks, figurative language remains peripheral, motivating dedicated evaluation frameworks for culturally grounded figurative understanding. 

\subsection{Improving Figurative Interpretation}

Prior work falls into two main directions: task-specific fine-tuning and
structured prompting. Supervised fine-tuning remains dominant:
\citet{yayavaram-etal-2024-bert} train a BERT-based model with objectives
capturing word cohesion and cross-lingual translation, yielding gains on idiom
detection, and \citet{pragmatic_sarcasm} show that fine-tuned encoder models
outperform decoder-only LLMs on multilingual sarcasm detection,
underscoring the difficulty of capturing implicit, culturally grounded meaning
through pretraining alone. A second line elicits figurative reasoning without
parameter updates:
\citet{prystawski2023psychologicallyinformedchainofthoughtpromptsmetaphor} use
psychologically informed chain-of-thought prompts for metaphor interpretation,
and \citet{lee-etal-2025-pragmatic} propose Pragmatic Metacognitive Prompting,
which scaffolds reasoning over implied meaning and literal--intended
discrepancies. Beyond these, a smaller body of work injects explicit knowledge or
human-inspired strategies: \citet{chakrabarty-etal-2022-rocket} model contextual
inference and compositional reasoning for idiom and simile interpretation in
narratives, showing that pretrained representations alone fall well short of
human performance.

\subsection{Cultural Adaptation and Cross-Domain Transfer}

A parallel literature adapts models to specific cultures, typically by fine-tuning
on culturally grounded data \cite{li2024culturellm, pham-etal-2025-cultureinstruct} or by
training separate adapters per culture \cite{adilazuarda-etal-2025-surveys}. For Arabic, the PalmX shared task
\cite{alwajih-etal-2025-palmx} has driven a set of cultural fine-tuning systems, and
\citet{almheiri-etal-2025-cross} show that cultural commonsense transfers
\emph{across} Arab countries under lightweight alignment.

In all of these, the target of adaptation is cultural knowledge itself, just as figurative benchmarks evaluate figurative competence on its own terms. We investigate instead whether supervision in either domain transfers to the other, and additionally whether it transfers across figurative forms, from poetry to both proverbs and idioms.

\section{Experimental Setup}

To examine the connection between culture and figurative language in LLMs, we design a controlled fine-tuning and evaluation framework that isolates the direction of transfer between the two domains. 
We design three complementary experiments to investigate the bidirectional transfer between cultural and figurative language understanding in Arabic LLMs.

\begin{enumerate}
    \setlength{\itemsep}{0pt}
    \setlength{\parskip}{0pt}
    \setlength{\topsep}{0pt}
    \item \textbf{The Impact of Cultural Fine-tuning: } We fine-tune on Arabic cultural datasets and evaluate the impact on LLM understanding of idioms and proverbs.
    \item \textbf{The Impact of Figurative Language Fine-tuning: } We fine-tune on Arabic poetry and proverbs and evaluate the impact on LLM understanding of idioms and proverbs. 
    \item \textbf{The Contribution of Language Content (Baseline):} We fine-tune on general Arabic data and evaluate on the same benchmarks, to test whether
transfer reflects cultural content specifically or Arabic exposure generally.
\end{enumerate}

\begin{table*}[ht]
\centering
\small

\begin{tabular}{p{3cm}  p{4cm} p{2cm} p{2.5cm} p{2cm}}
\toprule
\textbf{Dataset}  & \textbf{Description} & \textbf{Size} & \textbf{Coverage} & \textbf{Usage} \\
\midrule
\rowcolor{culturehighlight}
AraDiCE-Culture \cite{mousi2024aradicebenchmarksdialectalcultural} &
Multiple-choice cultural commonsense reasoning benchmark &
180 &
6 Arab countries &
Test \\
\rowcolor{culturehighlight}
ArabCulture \cite{sadallah2025commonsensereasoningarabculture} &
Cultural knowledge and practices across the Arab world &
3{,}482 &
13 Arab countries &
Train \\
\rowcolor{culturehighlight}
Palm \cite{alwajih-etal-2025-palm} &
Cultural commonsense and social reasoning in Arabic &
15.5k (train), 1.93k (test) &
22 Arab countries &
Train \\
\hline
\rowcolor{figurativehighlight}
FannOrFlop \cite{alghallabi2025fannflopmultigenremultiera} &
Poem--explanation pairs capturing poetic preference and aesthetic judgment across 14 genres, 12 eras &
6{,}984 &
Arabic poetry &
Train \\
\rowcolor{figurativehighlight}
Jawaher \cite{magdy-etal-2025-jawaher} &
Arabic proverb understanding and interpretation &
800 (train), 198 (test) &
20 Arabic varieties &
Train, test \\
\rowcolor{figurativehighlight}
Kinayat \cite{attia-etal-2026-beyond} &
Egyptian Arabic idiom--explanation pairs &
150 &
Egyptian Arabic &
Test \\
\hline
ArabicMMLU \cite{koto-etal-2024-arabicmmlu} &
Arabic language and grammar MCQs (control) & 980 &
Modern Standard Arabic & Train \\
\bottomrule
\end{tabular}
\caption{Datasets used for training and evaluation, covering figurative language and cultural knowledge across Arabic varieties and regions. \colorbox{culturehighlight}{Cultural} datasets (AraDiCE-Culture, ArabCulture, Palm) and \colorbox{figurativehighlight}{figurative language} datasets (FannOrFlop, Jawaher, Kinayat) are visually distinguished; ArabicMMLU serves as a control.}
\label{tab:datasets}
\end{table*}

\subsection{Datasets}\label{sec:datasets}

\paragraph{Training and Evaluation Data.}
Table~\ref{tab:datasets} summarizes all resources used; all are publicly
available. Figurative supervision comes from the Jawaher
\cite{magdy-etal-2025-jawaher} training split and FannOrFlop
\cite{alghallabi2025fannflopmultigenremultiera}, cultural supervision from
ArabCulture \cite{sadallah2025commonsensereasoningarabculture} and the
Palm \cite{alwajih-etal-2025-palm} training split. We evaluate cultural
understanding on AraDiCE
\cite{mousi2024aradicebenchmarksdialectalcultural} and figurative understanding on
the Jawaher test split and Kinayat
\cite{attia-etal-2026-beyond}.

\paragraph{Design Choices.}
First, because
Palm includes proverbs among its topic categories, we remove all
proverb-topic items, so that neither cultural condition carries figurative
supervision. Second, each condition is trained on a 1,000-sample subset, except Jawaher which only consists of 800 samples for training. Full-corpus training for FannOrFlop and ArabCulture is reported as an ablation in
Appendix~\ref{app:ablations}. Third, because every fine-tuning condition also constitutes exposure to Arabic
text, gains on any benchmark could reflect language adaptation rather than the
content of the training data. We therefore add a control condition trained on the
Arabic Language (General) and Arabic Language (Grammar) subsets of
ArabicMMLU \cite{koto-etal-2024-arabicmmlu}, 980 examples in total.

\paragraph{Data Preparation.}
All training data is cast as open-ended generation. For multiple-choice datasets,
the question serves as the input and the correct answer as the target, with
incorrect options discarded; for generative datasets, the input expression serves
as the input and its explanation or interpretation as the target. Prompt templates are given in Appendix~\ref{app:prompts}. For
Jawaher evaluation we use the distractors of
\citet{attia-etal-2026-beyond}; Kinayat and AraDiCE are used as
released. Jawaher and Kinayat have two options per item (50\% chance baseline) and AraDiCE three (33.3\%).

\subsection{Models and Evaluation}
We experiment with both multilingual and Arabic-focused instruction-tuned models. All models are fine-tuned using parameter-efficient Low-Rank Adaptation (LoRA) \cite{hu2021loralowrankadaptationlarge}. Fine-tuning setup is in Appendix~\ref{app:setup}.
Our multilingual models are LLaMA~3.1--8B~Instruct \cite{grattafiori2024llama3herdmodels} and Qwen~3--8B \cite{yang2025qwen3technicalreport}.
Our Arabic-centric models are Fanar-1-9B--Instruct \cite{fanarteam2025fanararabiccentricmultimodalgenerative} and ALLaM--7B--Instruct \cite{bari2025allam}.
All models are evaluated under identical training and inference conditions to ensure comparability. We evaluate models under the zero-shot setting, and each experiment is run three times with different random seeds for shuffling the multiple-choice options to account for sensitivity to answer ordering. All evaluations are run through the \texttt{lm-eval} \cite{eval-harness} framework, applied identically to base and fine-tuned models. Multiple-choice items are scored by comparing the log-likelihood the model assigns to each answer option, with an item counted correct when the gold option receives the highest log-likelihood. We report accuracy, averaged across the three runs.

\section{Results}\label{sec:results}

The one fine-tuning condition that produces a statistically reliable effect is poetry fine-tuning evaluated on idiom interpretation: FannOrFlop-finetuned models average $+2.33\%$ on Kinayat, the only pooled effect in our study whose 95\% confidence interval excludes zero ($[+0.39, +4.28]$, $p = 0.021$; Table~\ref{tab:clustered-aggregate}). Importantly, this is a within-figurative transfer effect rather than evidence of culture-to-figurative transfer. This effect is driven primarily by LLaMA-3.1-8B, whose largest single-seed gain of $+10.00\%$ is significant under an exact McNemar test ($p = 0.008$); the remaining two seeds gain $+8.00\%$ and $+2.67\%$, neither individually significant. Jawaher-finetuned models average $+1.83\%$ on Kinayat, in the same direction but not distinguishable from zero.

The most pronounced regressions occur under Palm fine-tuning, and these are the only negative effects with statistical support. ALLaM-7B fine-tuned on Palm degrades by $3.70\%$ on Jawaher, with one seed reaching significance ($-5.56\%$, $p = 0.013$), and Fanar-1-9B declines by $3.03\%$ on Jawaher under the same condition, likewise with one significant seed ($-3.54\%$, $p = 0.039$). Both regressions are consistent in sign across all three seeds. Other regressions we observe---LLaMA-3.1-8B fine-tuned on ArabCulture dropping $2.78\%$ on AraDiCE (Figure~\ref{fig:arabculture_heatmap}), and Qwen3-8B fine-tuned on Jawaher declining $2.36\%$ on Jawaher itself---fall within the range of seed-to-seed variation and should be treated as suggestive rather than established. Ablations on fine-tuning data size and hyperparameter choices were also conducted; full results are reported in Appendix~\ref{app:ablations}.

\begin{figure}[!ht]
    \centering
    \includegraphics[width=\columnwidth]{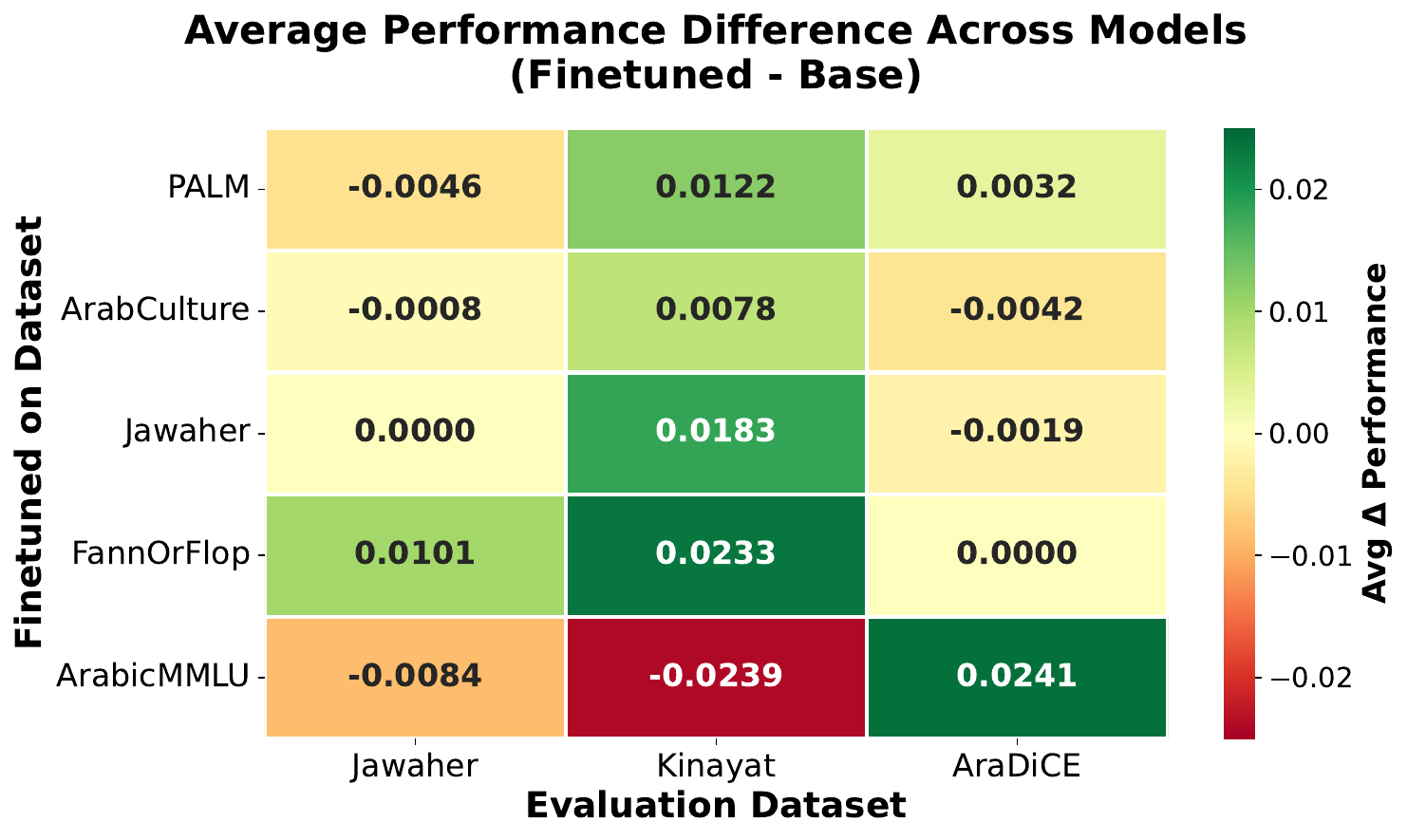}
    \caption{Average performance difference.}
    \label{fig:avg_heatmap}
    \vspace{-0.2in}
\end{figure}

\subsection{Transfer between Cultural and Figurative Language Fine-tuning}

Table~\ref{tab:average-results} and Figure~\ref{fig:avg_heatmap} summarize the performance differences (fine-tuned $-$ base) across all fine-tuning and evaluation dataset combinations, averaged over four models: ALLaM-7B, Fanar-1-9B, Qwen3-8B, and LLaMA-3.1-8B (See Appendix~\ref{app:more-results} for detailed results of individual runs and model heatmaps). Given the size of our evaluation sets (150--198 items), we quantify uncertainty for every reported effect using paired bootstrap confidence intervals over test items and exact McNemar tests, with a cluster bootstrap for effects pooled across models. Full per-seed intervals and test results appear in Appendix~\ref{app:significance}. Unless noted, differences discussed below are not statistically significant at $\alpha = 0.05$.

\begin{table}[ht]
\centering
\resizebox{\columnwidth}{!}{%
\begin{tabular}{ll|ccc}
\hline
& \textbf{Model} & \textbf{Jawaher} & \textbf{Kinayat} & \textbf{AraDiCE} \\
\hline
\multirow{4}{*}{Base} 
& ALLaM-7B-Instruct & 0.8990 & 0.8400 & 0.7537 \\
& Qwen3-8B & 0.8131 & 0.6800 & 0.5037 \\
& Fanar-1-9B-Instruct & 0.8906 & 0.7667 & 0.6870 \\
& Llama-3.1-8B-Instruct & 0.6886 & 0.5956 & 0.5204 \\
\hline
\multirow{4}{*}{\makecell{Fine-tuned on\\ Palm Subset}} 
& ALLaM-7B-Instruct & 0.8620 & 0.8333 & 0.7167 \\
& Qwen3-8B & 0.8300 &	0.7178 & 0.5204 \\
& Fanar-1-9B-Instruct & 0.8603 & 0.7467 & 0.6926 \\
& Llama-3.1-8B-Instruct & 0.7205 & 0.6333 & 0.5481 \\
\hline
\multirow{4}{*}{\makecell{Fine-tuned on\\ ArabCulture\\ subset}} 
& ALLaM-7B-Instruct & 0.8906 & 0.8267 & 0.7500 \\
& Qwen3-8B & 0.8148 & 0.7133 & 0.5185 \\
& Fanar-1-9B-Instruct & 0.8805 & 0.7533 & 0.6870 \\
& Llama-3.1-8B-Instruct & 0.7020 & 0.6200 & 0.4926 \\
\hline
\multirow{4}{*}{\makecell{Fine-tuned on\\ Jawaher}} 
& ALLaM-7B-Instruct & 0.8990 & 0.8533 & 0.7407 \\
& Qwen3-8B & 0.7896 & 0.6756 & 0.5130 \\
& Fanar-1-9B-Instruct & 0.8906 & 0.7489 & 0.6907 \\
& Llama-3.1-8B-Instruct & 0.7121 & 0.6778 & 0.5130 \\
\hline
\multirow{4}{*}{\makecell{Fine-tuned on\\FannOrFlop}}
& ALLaM-7B-Instruct & 0.8889 & 0.8289 & 0.7352 \\
& Qwen3-8B & 0.8333 & 0.7000 & 0.5111 \\
& Fanar-1-9B-Instruct & 0.8973 & 0.7822 & 0.6944 \\
& Llama-3.1-8B-Instruct & 0.7121 & 0.6644 & 0.5241 \\
\hline
\multirow{4}{*}{\makecell{Fine-tuned on\\ ArabicMMLU\\ (control)}}
& ALLaM-7B-Instruct & 0.8906 & 0.7897 & 0.7833 \\
& Qwen3-8B & 0.7795 & 0.6369 & 0.5389 \\
& Fanar-1-9B-Instruct & 0.8838 & 0.7436 & 0.6926 \\
& Llama-3.1-8B-Instruct & 0.7037 & 0.6164 & 0.5463 \\
\hline
\end{tabular}%
}
\caption{Average evaluation results across three runs on Jawaher, Kinayat, and AraDiCE datasets. Results show accuracy scores ($\uparrow$) for base models and models fine-tuned on different subsets.}
\label{tab:average-results}
\end{table}

\paragraph{Effect of Cultural Fine-tuning on Figurative Language Benchmarks.}
Fine-tuning on ArabCulture and Palm---both cultural datasets---does not reliably improve performance on the figurative language benchmarks Jawaher and Kinayat. ArabCulture produces near-zero change on Jawaher ($-0.08\%$, $p = 0.92$) and a small positive point estimate on Kinayat ($+0.78\%$, $p = 0.44$), while Palm yields $+1.22\%$ on Kinayat ($p = 0.29$) and $-0.46\%$ on Jawaher ($p = 0.58$). All four confidence intervals span zero. Cultural knowledge alone therefore appears to provide no measurable transfer to tasks requiring understanding of idioms and proverbs.

The pooled estimates, however, average over sharply divergent per-model behavior 
(Figure~\ref{fig:palm_heatmap}). Under Palm fine-tuning, ALLaM-7B and Fanar-1-9B both regress across all three evaluation datasets, and as noted above their Jawaher regressions are the only negative effects in our study with statistical support. Qwen3-8B and LLaMA-3.1-8B move in the opposite direction, with LLaMA-3.1-8B posting positive point estimates of $+3.20\%$, $+3.78\%$, and $+2.78\%$ on Jawaher, Kinayat, and AraDiCE respectively; none of the underlying per-seed comparisons is individually significant, so we read this as a consistent sign pattern rather than a demonstrated gain. Qwen3-8B records one significant seed under Palm fine-tuning ($+4.67\%$ on Kinayat, $p = 0.016$). 
That the two Arabic-centric models and the two multilingual models respond in opposite directions, with significance on both sides, is the clearest evidence in our results that the effect of cultural fine-tuning is model-dependent.

\paragraph{Effect of Figurative Language Fine-tuning on AraDiCE.}
Fine-tuning on Jawaher (proverbs) and FannOrFlop (poetry) produces no detectable improvements on AraDiCE. The pooled effects are $-0.19\%$ ($p = 0.81$) and $+0.00\%$ ($p = 0.99$) respectively, both tightly centered on zero. At the model level 
(Figures~\ref{fig:fannorflop_heatmap} and~\ref{fig:jawaher_heatmap}), Qwen3-8B and LLaMA-3.1-8B show small positive point estimates while ALLaM-7B shows negative ones; the ALLaM-7B average of $-1.85\%$ under FannOrFlop rests on a single significant seed ($-5.00\%$, $p = 0.022$) against seeds of $+1.11\%$ and $-1.67\%$, so the size of this regression should be read with caution even though its direction is supported. Overall, fine-tuning on proverb and poetry datasets does not improve performance on more general Arabic cultural commonsense tasks.

\paragraph{Control Fine-tuning on ArabicMMLU.}
The ArabicMMLU control does not reproduce the pattern observed under content-matched fine-tuning. Averaged across models, it yields $-2.39\%$ on Kinayat and $-0.84\%$ on Jawaher (Figure~\ref{fig:arabicmmlu_heatmap}), with three of four models regressing on each. On Kinayat the control thus moves in the opposite direction from poetry fine-tuning, a divergence of nearly five percentage points, indicating that the reliable gain reported above is not explained by adaptation to the fine-tuning format. ALLaM-7B records the single largest regression in our study under this condition ($-5.03\%$ on Kinayat), consistent with the broader pattern that the Arabic-centric models respond poorly to fine-tuning on data whose content they have likely already absorbed.

The control does, however, produce the only unanimous effect we observe: all four models improve on AraDiCE, averaging $+2.41\%$. AraDiCE does not move under every other condition, including both cultural sets, so the one condition that moves it targets general language ability rather than cultural knowledge. ArabicMMLU is not culturally neutral: as natively produced rather than translated Arabic, it carries cultural content incidentally, so our contrast is between curated and incidental cultural exposure rather than cultural and non-cultural data. With that qualification, we hypothesize that AraDiCE performance is limited less by cultural knowledge than by general Arabic competence, inverting the assumption that motivates cultural fine-tuning for cultural commonsense benchmarks.

\subsection{General Observations}

Several patterns emerge across fine-tuning conditions. First, the Kinayat benchmark is the most responsive to fine-tuning and is the only benchmark on which any fine-tuning condition produces a reliable aggregate gain. It also shows the widest spread of individual outcomes, ranging from LLaMA-3.1-8B's $+8.22\%$ under Jawaher fine-tuning to small regressions for ALLaM-7B and Fanar-1-9B under the same conditions.

Second, the Arabic-centric models benefit less from fine-tuning. ALLaM-7B and Fanar-1-9B start from substantially higher baselines and record negative deltas in the majority of fine-tuning and evaluation combinations, including the two significant regressions on Jawaher under Palm fine-tuning discussed above. Their consistently higher baselines, combined with the direction of these effects, suggest that these models already capture much of the cultural and figurative knowledge present in the fine-tuning corpora. The multilingual models Qwen3-8B and LLaMA-3.1-8B show positive point estimates more often, and account for the two significant positive cells in our results, though most of their individual gains are not distinguishable from zero.

Third, and most consistently, the average performance differences across all fine-tuning datasets (Figure~\ref{fig:avg_heatmap}) are uniformly small in magnitude (ranging from $-0.46\%$ to $+2.33\%$), and mostly statistically indistinguishable from zero. Cross-task transfer between cultural knowledge and figurative 
language understanding in Arabic is therefore limited, model-dependent, and---with the single exception of poetry fine-tuning on idiom interpretation---not reliably positive. The one reliable effect we observe is not reproduced by our 
ArabicMMLU control, which regresses on Kinayat where poetry fine-tuning gains, indicating that the improvement stems from the figurative content of the poetry data rather than from adaptation to the fine-tuning format.

\section{Error Analysis}

\subsection{Figurative Language and Cultural Understanding}

We analyze the impact of figurative language fine-tuning (Jawaher proverbs and FannOrFlop poetry) on culturally grounded QA across models, countries, and topic categories. Improvements (185) and regressions (189) are nearly balanced, yielding a slight net degradation ($-4$) and negligible average change ($-0.11\%$), consistent with the null pooled effects reported in Section~\ref{sec:results}. The analysis below therefore characterizes the item-level churn underlying that null result rather than a net gain or loss; because counts aggregate over models and seeds, and several categories contain few items, the patterns are descriptive and we do not test them individually.

\paragraph{Model-Level Trends} Table~\ref{tab:model_analysis} summarizes performance changes across models and datasets. 
Fanar and Qwen post small positive point estimates under both fine-tuning sets, while ALLaM shows the largest degradation, particularly under poetry ($-1.87\%$); this last figure rests on a single significant seed ($-5.00\%$, $p = 0.022$) against seeds of $+1.11\%$ and $-1.67\%$, so its direction is better supported than its magnitude. LLaMA is mixed across conditions.

\begin{table}[ht]
\centering
\small
\resizebox{\columnwidth}{!}{%
\begin{tabular}{llcccc}
\toprule
\textbf{Model} & \textbf{Train Set} & \textbf{Avg $\Delta$ \%} & \textbf{Impr} & \textbf{Regr} & \textbf{Net} \\
\midrule
\multirow{2}{*}{ALLaM} & Jawaher    & -1.30 & 21 & 28 & -7  \\
                       & FannOrFlop & -1.87 & 15 & 25 & -10 \\
\hline
\multirow{2}{*}{Fanar} & Jawaher    & +0.37 & 22 & 20 & +2  \\
                       & FannOrFlop & +0.73 & 23 & 19 & +4  \\
\hline
\multirow{2}{*}{Llama} & Jawaher    & -0.77 & 38 & 42 & -4  \\
                       & FannOrFlop & +0.37 & 43 & 41 & +2  \\
\hline 
\multirow{2}{*}{Qwen}  & Jawaher    & +0.87 & 10 & 5  & +5  \\
                       & FannOrFlop & +0.70 & 13 & 9  & +4  \\
\bottomrule
\end{tabular}
}
\caption{Aggregate performance changes on culture by model and fine-tuning dataset.}
\label{tab:model_analysis}
\end{table}

\paragraph{Geographic Variation} Performance changes vary across countries (Table~\ref{tab:country_topic_analysis}). These counts reflect aggregated improvement and regression instances across all models and fine-tuning datasets, such that a single question may contribute multiple counts if it consistently improved or regressed across configurations. Jordan accumulates the strongest net improvement ($+32$), while most other countries experience net regressions, with the largest degradations observed for Syria ($-18$) and Egypt ($-12$). Since the 
aggregate effect is null, this dispersion indicates that fine-tuning redistributes 
performance across cultural contexts rather than improving it uniformly.

\begin{table}[h]
\centering
\small
\begin{tabular}{lccc}
\toprule
 & \textbf{Impr} & \textbf{Regr} & \textbf{Net} \\
\midrule
\multicolumn{4}{l}{\textit{By Country}} \\
\midrule
Jordan     & 50 & 18 & $+32$ \\
Lebanon    & 30 & 29 & $+1$  \\
Qatar      & 26 & 28 & $-2$  \\
Palestine  & 30 & 35 & $-5$  \\
Egypt      & 22 & 34 & $-12$ \\
Syria      & 27 & 45 & $-18$ \\
\midrule
\multicolumn{4}{l}{\textit{By Topic}} \\
\midrule
Food/Cuisine          & 18 & 3  & $+15$ \\
Traditional Games     & 18 & 5  & $+13$ \\
Other                 & 66 & 65 & $+1$  \\
Holidays/Occasions    & 40 & 46 & $-6$  \\
History/Civilization  & 3  & 11 & $-8$  \\
Religion              & 2  & 10 & $-8$  \\
Traditional Clothing  & 17 & 31 & $-14$ \\
\bottomrule
\end{tabular}
\caption{Performance changes on cultural evaluation by country and topic category, aggregated across all models and fine-tuning configurations.}
\label{tab:country_topic_analysis}
\end{table}

\paragraph{Topic-Level Analysis} We further analyze performance across cultural topics 
(Table~\ref{tab:country_topic_analysis}). The strongest net gains appear in 
\textit{Food/Cuisine} ($+15$) and \textit{Traditional Games} ($+13$), while the largest net 
losses occur in \textit{Traditional Clothing} ($-14$), \textit{History/Civilization} 
($-8$), and \textit{Religion} ($-8$); the latter two rest on few items and should be read 
with caution. At the question level, the most improved items cluster around popular 
culture: traditional games, sweets, marriage customs, and celebrations such as Ramadan. 
Regressions concentrate on historical and political content---battles, the Islamic conquest 
of Damascus, independence dates---as well as traditional clothing and infrastructural 
knowledge (Table~\ref{tab:aradice_top_improved_regressed}). This suggests that where 
figurative fine-tuning does shift cultural predictions, it tends to favor 
\textit{culturally embedded, experiential knowledge} over \textit{historically grounded or 
domain-specific factual knowledge}---a topical sensitivity that goes beyond model or dialect differences.

\subsection{Culture and Poetry as Signals for Figurative Language}

The positive effect of poetry fine-tuning on idiom interpretation is noteworthy. While one might intuitively sense a connection---poetry, idioms, and proverbs all operate in non-literal registers---this relationship is rarely foregrounded in NLP research, which tends to treat these as distinct figurative categories with separate datasets, tasks, and benchmarks. Poetry draws on imagery, meter, and rhetorical devices, whereas idioms and proverbs rely on fixed, conventionalized meanings; the overlap is conceptual rather than structural. The results here suggest, however, that fine-tuning on poetry develops a broader figurative competence---a sensitivity to non-literal meaning that is not tied to any one figurative type but transfers across them. We note that the same effect on proverbs points in the same direction but does not reach significance ($+1.01\%$, $p = 0.23$), so the evidence for such competence rests on the poetry-to-idiom pairing rather than on transfer to figurative language generally.

\paragraph{Overall Fine-Tuning Effect} Table~\ref{tab:overall_summary} compares fine-tuning outcomes across both figurative tasks. Cultural fine-tuning shifts idiom accuracy by +1.00\% and proverb accuracy by -0.27\%, neither distinguishable from zero, whereas poetry fine-tuning is positive on both and reliably so on idioms (+2.33\% vs. +1.01\%). Effects are smaller on proverbs under both conditions, which may reflect the greater dialectal diversity of the proverb set. Per-model breakdowns are given in Tables~\ref{tab:per_file_idioms_cultural_poetry} and~\ref{tab:per_file_proverbs_cultural_poetry} (Appendix~\ref{Appendix-error}).

\begin{table}[h]
\centering
\small
\renewcommand{\arraystretch}{1.2}
\begin{tabular}{lrrrr}
\hline
\textbf{Task} & \textbf{Impr} & \textbf{Regr} & \textbf{Net} & \textbf{Avg $\Delta\%$} \\
\hline
\multicolumn{5}{l}{\textit{Cultural Fine-tuning}} \\
\hline
Idioms   & 195 & 159 & $+36$ & $+1.00$ \\
Proverbs & 149 & 162 & $-13$ & $-0.27$ \\
\hline
\multicolumn{5}{l}{\textit{Poetry Fine-tuning}} \\
\hline
Idioms   & 100 & 58  & $+42$ & $+2.33$ \\
Proverbs & 97  & 73  & $+24$ & $+1.01$ \\
\hline
\end{tabular}
\caption{Overall fine-tuning effect on idiom and proverb interpretation, aggregated across all models. Cultural fine-tuning uses ArabCulture and Palm; Poetry fine-tuning uses FannOrFlop.}
\label{tab:overall_summary}
\end{table}

\vspace{-10pt}

\paragraph{Model-Level Trends} Table~\ref{tab:aggregate} aggregates results by model and evaluation dataset for both tasks. The pattern is consistent across fine-tuning conditions: ALLaM and Fanar regress while LLaMA and Qwen improve, with the magnitude of harm greater on proverbs. ALLaM on Palm proverbs loses $-3.70\%$ (net $-22$) against only $-0.67\%$ (net $-3$) for idioms, and its worst single run is a $-5.56\%$ drop on proverbs (Palm, seed~21, $p = 0.013$). This amplified degradation for the stronger Arabic-centric models is consistent with their having already internalized proverb knowledge during pre-training on large Arabic web corpora, so that fine-tuning on a curated, narrower corpus introduces noise relative to that broader prior. LLaMA and Qwen gain on both tasks but less on proverbs; Qwen accumulates only 4 regressions on idioms against 18 on proverbs, though its net remains positive. Poetry fine-tuning broadly mirrors this pattern, with one exception: Fanar now benefits on both idioms ($+1.56\%$, net $+7$) and proverbs ($+0.67\%$, net $+4$), whereas cultural fine-tuning yielded only losses. Poetry fine-tuning produces the largest single-seed improvement in LLaMA's idiom performance ($+10.00\%$, seed~21, $p = 0.008$), with a net improvement of $+31$ predictions across the three seeds.

\begin{table}[h]
\centering
\small
\resizebox{\columnwidth}{!}{%
\begin{tabular}{llrr@{\hspace{1.5em}}rr}
\hline
 & & \multicolumn{2}{c}{\textbf{Idioms}} & \multicolumn{2}{c}{\textbf{Proverbs}} \\
\cline{3-4} \cline{5-6}
\textbf{Model} & \textbf{Train Set} & \textbf{Avg $\Delta$\%} & \textbf{Net} & \textbf{Avg $\Delta$\%} & \textbf{Net} \\
\hline
\multirow{3}{*}{ALLaM} 
  & \cellcolor{culturehighlight}ArabCulture & \cellcolor{culturehighlight}$-1.33$ & \cellcolor{culturehighlight}$-6$  & \cellcolor{culturehighlight}$-0.84$ & \cellcolor{culturehighlight}$-5$  \\
  & \cellcolor{culturehighlight}Palm        & \cellcolor{culturehighlight}$-0.67$ & \cellcolor{culturehighlight}$-3$  & \cellcolor{culturehighlight}$-3.70$ & \cellcolor{culturehighlight}$-22$ \\
  & \cellcolor{poetryhighlight}Poetry      & \cellcolor{poetryhighlight}$-1.11$ & \cellcolor{poetryhighlight}$-5$  & \cellcolor{poetryhighlight}$-1.01$ & \cellcolor{poetryhighlight}$-6$  \\
\hline
\multirow{3}{*}{Fanar} 
  & \cellcolor{culturehighlight}ArabCulture & \cellcolor{culturehighlight}$-1.33$ & \cellcolor{culturehighlight}$-6$  & \cellcolor{culturehighlight}$-1.01$ & \cellcolor{culturehighlight}$-6$  \\
  & \cellcolor{culturehighlight}Palm        & \cellcolor{culturehighlight}$-2.00$ & \cellcolor{culturehighlight}$-9$  & \cellcolor{culturehighlight}$-3.03$ & \cellcolor{culturehighlight}$-18$ \\
  & \cellcolor{poetryhighlight}Poetry      & \cellcolor{poetryhighlight}$+1.56$ & \cellcolor{poetryhighlight}$+7$  & \cellcolor{poetryhighlight}$+0.67$ & \cellcolor{poetryhighlight}$+4$  \\
\hline
\multirow{3}{*}{LLaMA} 
  & \cellcolor{culturehighlight}ArabCulture & \cellcolor{culturehighlight}$+2.44$ & \cellcolor{culturehighlight}$+11$ & \cellcolor{culturehighlight}$+1.35$ & \cellcolor{culturehighlight}$+8$  \\
  & \cellcolor{culturehighlight}Palm        & \cellcolor{culturehighlight}$+3.78$ & \cellcolor{culturehighlight}$+17$ & \cellcolor{culturehighlight}$+3.20$ & \cellcolor{culturehighlight}$+19$ \\
  & \cellcolor{poetryhighlight}Poetry      & \cellcolor{poetryhighlight}$+6.89$ & \cellcolor{poetryhighlight}$+31$ & \cellcolor{poetryhighlight}$+2.36$ & \cellcolor{poetryhighlight}$+14$ \\
\hline
\multirow{3}{*}{Qwen}  
  & \cellcolor{culturehighlight}ArabCulture & \cellcolor{culturehighlight}$+3.33$ & \cellcolor{culturehighlight}$+15$ & \cellcolor{culturehighlight}$+0.17$ & \cellcolor{culturehighlight}$+1$  \\
  & \cellcolor{culturehighlight}Palm        & \cellcolor{culturehighlight}$+3.78$ & \cellcolor{culturehighlight}$+17$ & \cellcolor{culturehighlight}$+1.68$ & \cellcolor{culturehighlight}$+10$ \\
  & \cellcolor{poetryhighlight}Poetry      & \cellcolor{poetryhighlight}$+2.02$ & \cellcolor{poetryhighlight}$+9$  & \cellcolor{poetryhighlight}$+2.00$ & \cellcolor{poetryhighlight}$+12$ \\
\hline
\end{tabular}
}
\caption{Aggregate fine-tuning results by model and fine-tuning dataset for both idiom and proverb interpretation. \colorbox{culturehighlight}{Cultural} fine-tuning datasets (ArabCulture, Palm) and \colorbox{poetryhighlight}{Poetry} fine-tuning (FannOrFlop) are visually distinguished. Net refers to the total net improved predictions across all three seeds.}
\label{tab:aggregate}
\end{table}

\section{Conclusion}

This work examined whether fine-tuning on cultural and figurative language can induce transfer across the two domains. Across four models and six datasets spanning diverse Arabic dialects and regions, we find that this transfer is limited, inconsistent, and highly model-dependent. When cultural datasets serve as the training signal, pooled gains on figurative benchmarks are small and unreliable ($+0.78\%$ for ArabCulture and $+1.22\%$ for Palm on Kinayat), and the underlying per-model effects are polarized in sign; Palm fine-tuning produces the only statistically supported regressions in our study, lowering the accuracy of both Arabic-centric models on Jawaher. In the opposite direction, figurative fine-tuning produces no detectable change on cultural benchmarks.

The one reliable effect runs within the figurative domain rather than across the 
cultural--figurative divide: poetry fine-tuning on FannOrFlop improves idiom interpretation ($+2.33\%$, $p = 0.021$), and our ArabicMMLU control regresses on the same benchmark, indicating that this gain stems from figurative content rather than additional language exposure. A consistent divide also emerges between model families: the Arabic-centric models start from higher baselines and account for both supported regressions, suggesting they have already internalized much of the relevant knowledge during pre-training, whereas the multilingual models show greater headroom. Error analysis further reveals that fine-tuning tends to favor culturally embedded, experiential knowledge (such as food, 
customs, and celebrations) over historically grounded or domain-specific factual knowledge. Given our small evaluation set sizes, these findings are best read as a constraint on what current Arabic resources can support rather than evidence that no such relationship exists.

\section*{Limitations}

Our work is subject to several limitations. First, our fine-tuning experiments are limited to LoRA due to constraints in computational resources. While LoRA has proven effective in many settings, full fine-tuning or alternative adaptation methods may yield different results, and we were unable to explore this space. Second, hyperparameter search was limited in scope; factors such as learning rate, rank, and training duration were not exhaustively explored, and it is possible that more carefully tuned configurations would produce stronger or more consistent results. Third, our experiments are restricted to models in the 7--9 billion parameter range. Larger models may exhibit different sensitivity to cultural and poetic fine-tuning data, and our findings may not generalize to models at greater scales. We leave broader exploration of model sizes, fine-tuning strategies, and hyperparameter configurations to future work.

A limitation of our cross-domain analysis is that the two transfer directions are not directly comparable. They involve different source datasets, different target benchmarks, and different task formulations, so the observed difference in magnitude between directions cannot be cleanly attributed to a directional asymmetry in the underlying relationship — it may instead reflect variation in dataset difficulty, domain specificity, or evaluation sensitivity. We therefore avoid drawing strong conclusions about which direction of transfer is inherently stronger. We also find that neither domain reliably generalizes to the other across all models and conditions, which further limits the generality of any cross-domain claims we can make from these experiments.

\bibliography{custom}

\appendix
\section{Prompt Templates}\label{app:prompts} 

Figure~\ref{fig:mcq-prompt} shows the prompt used for the MCQ understanding task, which presents the model with a question and a set of candidate answers and elicits a single-letter selection. Figure~\ref{fig:prompt_aradice} shows the corresponding template for AraDiCE, which follows the same structure with three candidate options. Figures~\ref{fig:fannorflop-prompt}, \ref{fig:jawaher-prompt}, and \ref{fig:arabculture-prompt} present the prompts used for fine-tuning on the FannOrFlop poetry explanation, Jawaher proverb explanation, and ArabCulture open-ended completion tasks, respectively. Each of these prompts pairs a short task-specific preamble with the input field corresponding to that task (a poem, a proverb, or a cultural prompt) and instructs the model to produce a free-form explanation or completion. Finally, Figure~\ref{fig:palm-prompt} shows the prompt used for fine-tuning on the Palm instruction-following task; in contrast to the other tasks, no task-specific preamble is used, and the question field alone carries the full context required for the model to generate a response.

\begin{figure}[!htb]
\centering
\begin{tcolorbox}[
  colback=gray!10!white, colframe=black,
  boxrule=0.7pt, arc=2pt,
  left=6pt, right=6pt, top=6pt, bottom=6pt,
  width=0.95\columnwidth]
You are tasked with selecting the correct explanation for the following proverb.

Choose the correct explanation from the options provided. Only output the letter corresponding to the correct answer and nothing else.  \\

\textbf{Proverb:} \texttt{[PROVERB]} \\

\textbf{Options:} A. \texttt{[OPTION 1]} \\
\hspace*{2.7em} B. \texttt{[OPTION 2]} \\

\textbf{Answer:}
\end{tcolorbox}

\caption{Prompt used for the MCQ understanding task.}
\label{fig:mcq-prompt}
\end{figure}

\begin{figure}[!htb]
\centering
\begin{tcolorbox}[
  colback=gray!10!white, colframe=black,
  boxrule=0.7pt, arc=2pt,
  left=6pt, right=6pt, top=6pt, bottom=6pt,
  width=0.95\columnwidth]
You are tasked with selecting the correct answer to the following question.
Choose the correct answer from the options provided. Only output the letter corresponding to the correct answer and nothing else.  \\
\textbf{Question:} \texttt{[QUESTION]} \\
\textbf{Options:} A. \texttt{[OPTION 1]} \\
\hspace*{2.7em} B. \texttt{[OPTION 2]} \\
\hspace*{2.7em} C. \texttt{[OPTION 3]} \\
\textbf{Answer:}
\end{tcolorbox}
\caption{Prompt template used for AraDiCE-Culture evaluation.}
\label{fig:prompt_aradice}
\end{figure}

\begin{figure}[!htb]
\centering
\begin{tcolorbox}[
  colback=gray!10!white, colframe=black,
  boxrule=0.7pt, arc=2pt,
  left=6pt, right=6pt, top=6pt, bottom=6pt,
  width=0.95\columnwidth]
Explain the following poetry verse. \\
\textbf{Verse:} \texttt{[VERSE]} \\
\textbf{Explanation:} \texttt{[EXPLANATION]}
\end{tcolorbox}
\caption{Prompt used for fine-tuning on FannOrFlop poetry explanation.}
\label{fig:fannorflop-prompt}
\end{figure}

\begin{figure}[!htb]
\centering
\begin{tcolorbox}[
  colback=gray!10!white, colframe=black,
  boxrule=0.7pt, arc=2pt,
  left=6pt, right=6pt, top=6pt, bottom=6pt,
  width=0.95\columnwidth]
Explain the following Arabic proverb. \\
\textbf{Proverb:} \texttt{[PROVERB]} \\
\textbf{Explanation:} \texttt{[AR\_EXPLANATION]}
\end{tcolorbox}
\caption{Prompt used for fine-tuning on the Jawaher proverb explanation task.}
\label{fig:jawaher-prompt}
\end{figure}

\begin{figure}[!htb]
\centering
\begin{tcolorbox}[
  colback=gray!10!white, colframe=black,
  boxrule=0.7pt, arc=2pt,
  left=6pt, right=6pt, top=6pt, bottom=6pt,
  width=0.95\columnwidth]
Based on the following cultural context, provide an appropriate completion. \\
\textbf{Scenario:} \texttt{[SCENARIO]} \\
\textbf{Answer:} \texttt{[EXPECTED\_ANSWER]}
\end{tcolorbox}
\caption{Prompt used for fine-tuning on ArabCulture open-ended completion.}
\label{fig:arabculture-prompt}
\end{figure}
 
\begin{figure}[!htb]
\centering
\begin{tcolorbox}[
  colback=gray!10!white, colframe=black,
  boxrule=0.7pt, arc=2pt,
  left=6pt, right=6pt, top=6pt, bottom=6pt,
  width=0.95\columnwidth]
\texttt{[INSTRUCTION]} \\[6pt]
\texttt{[OUTPUT]}
\end{tcolorbox}
\caption{Prompt used for fine-tuning on the Palm instruction-following task. No task-specific preamble is used; the instruction field carries full context.}
\label{fig:palm-prompt}
\end{figure}

\section{Fine-tuning Setup}\label{app:setup}
We apply LoRA to the attention and feed-forward projection layers with rank $r=4$, scaling factor $\alpha=8$, and dropout $0.1$.
Fine-tuning is performed for $3$ epochs using a learning rate of $5e-5$ and a batch size of $1$.
A validation set is used to monitor convergence. This setup enables efficient adaptation while keeping the base model parameters frozen. All models were fine-tuned on a single NVIDIA RTX 5000 Ada Generation GPU. Additional hyperparameters were also explored.

\section{Additional Results}\label{app:more-results}
\begin{figure}[!ht]
    \centering
    \includegraphics[width=\columnwidth]{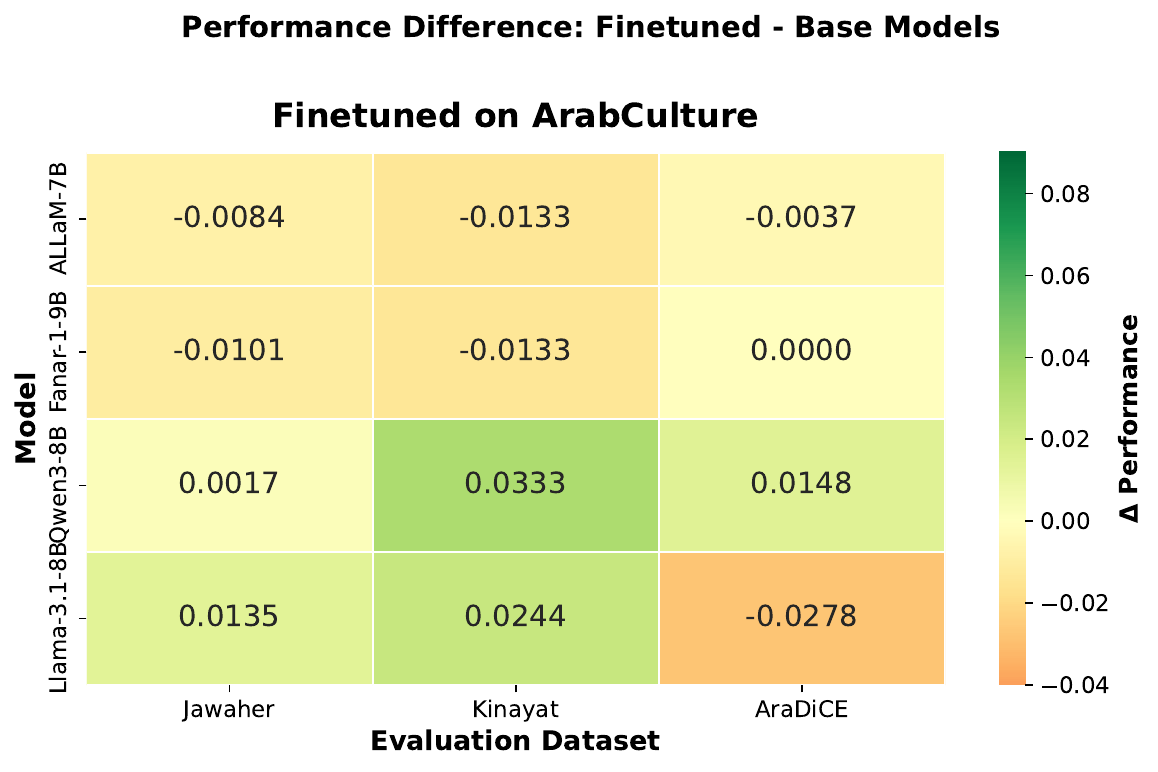}
    \caption{Performance difference on different datasets for models fine-tuned on the ArabCulture dataset (diff = fine-tuned accuracy - base accuracy).}
    \label{fig:arabculture_heatmap}
    \vspace{-0.2in}
\end{figure}

\begin{figure}[!ht]
    \centering
    \includegraphics[width=\columnwidth]{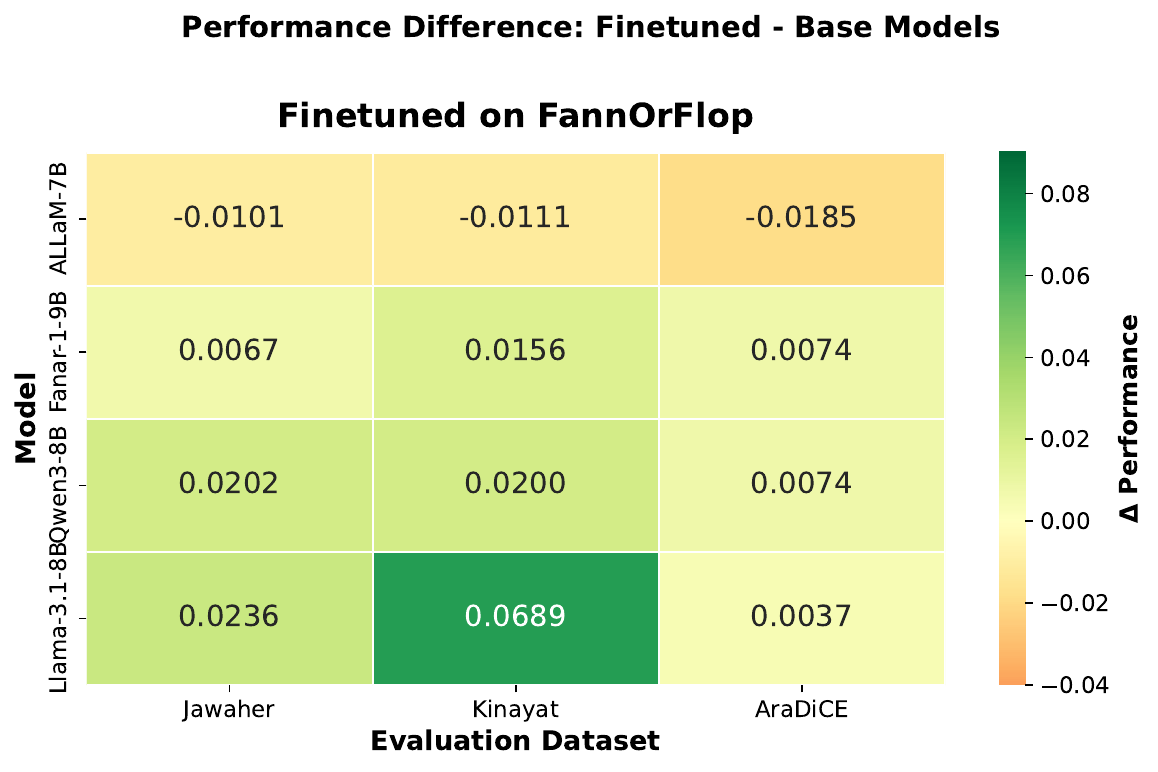}
    \caption{Performance difference on different datasets for models fine-tuned on the FannOrFlop dataset (diff = fine-tuned accuracy - base accuracy).}
    \label{fig:fannorflop_heatmap}
    \vspace{-0.2in}
\end{figure}

\begin{figure}[!ht]
    \centering
    \includegraphics[width=\columnwidth]{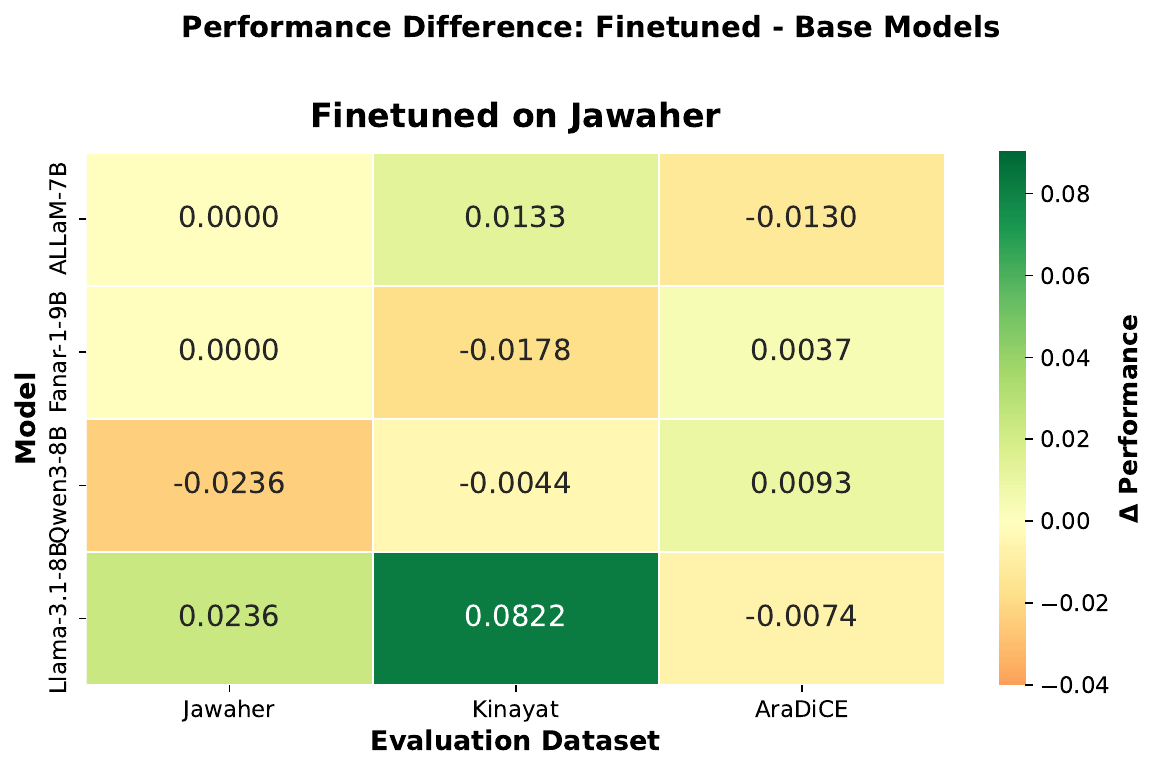}
    \caption{Performance difference on different datasets for models fine-tuned on the Jawaher dataset (diff = fine-tuned accuracy - base accuracy).}
    \label{fig:jawaher_heatmap}
    \vspace{-0.2in}
\end{figure}

\begin{figure}[!ht]
    \centering
    \includegraphics[width=\columnwidth]{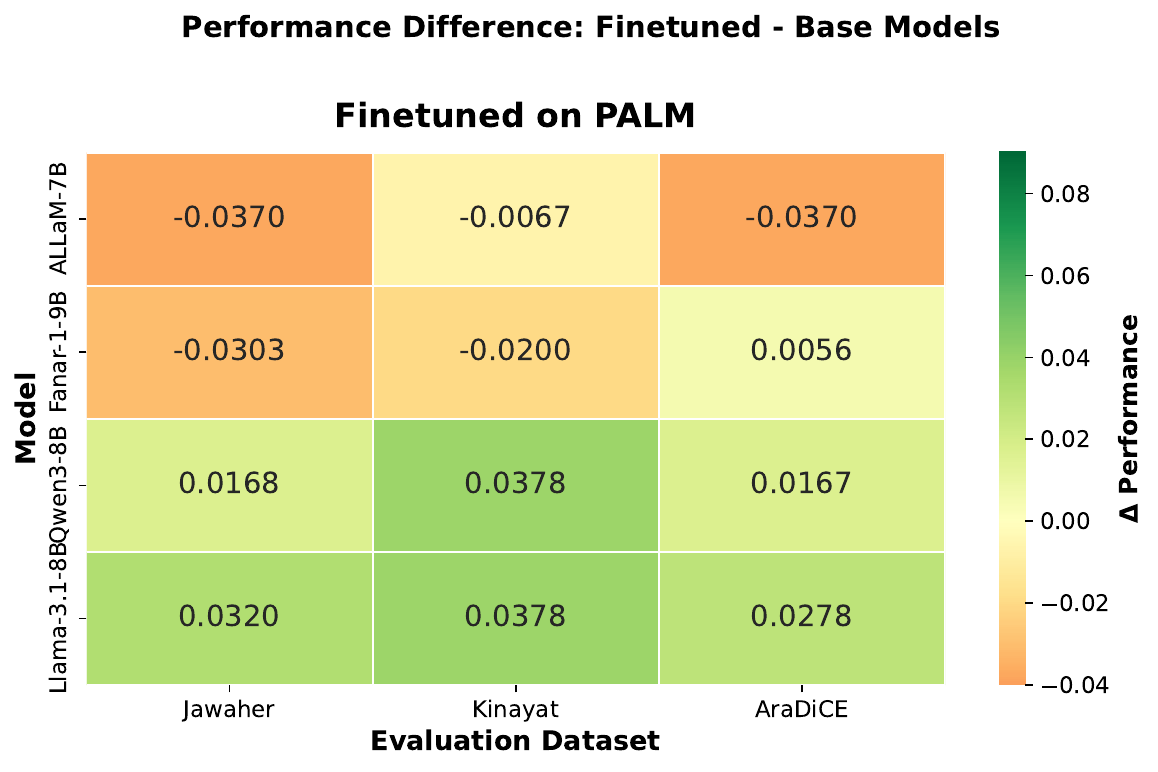}
    \caption{Performance difference on different datasets for models fine-tuned on the Palm dataset (diff = fine-tuned accuracy - base accuracy).}
    \label{fig:palm_heatmap}
    \vspace{-0.2in}
\end{figure}

Table~\ref{tab:zero-shot-all-runs} presents evaluation results across three runs with different random seeds on the Jawaher, Kinayat, and AraDiCE datasets, reporting accuracy scores (↑) for both the base models and the models fine-tuned on different subsets; reporting multiple runs allows us to assess the stability of the observed trends under random initialization. Figures~\ref{fig:arabculture_heatmap}, \ref{fig:fannorflop_heatmap}, \ref{fig:jawaher_heatmap}, and \ref{fig:palm_heatmap} visualize, as heatmaps, the per-dataset performance differences (diff = fine-tuned accuracy $-$ base accuracy) for models fine-tuned on the ArabCulture, FannOrFlop, Jawaher, and Palm datasets, respectively. Each heatmap reports the change in accuracy on the held-out evaluation datasets.

\begin{table*}[t]
\centering
\tiny
\begin{tabular}{llccccccccc}
\toprule
& & \multicolumn{3}{c}{\textbf{Run 1 (seed=0)}} & \multicolumn{3}{c}{\textbf{Run 2 (seed=42)}} & \multicolumn{3}{c}{\textbf{Run 3 (seed=21)}} \\
\cmidrule(lr){3-5} \cmidrule(lr){6-8} \cmidrule(lr){9-11}
\textbf{} & \textbf{Model} & \textbf{Jawaher} & \textbf{Kinayat} & \textbf{AraDiCE} & \textbf{Jawaher} & \textbf{Kinayat} & \textbf{AraDiCE} & \textbf{Jawaher} & \textbf{Kinayat} & \textbf{AraDiCE} \\
\midrule
\multirow{4}{*}{Base} 
& ALLaM-7B-Instruct & 0.8889 & 0.8600 & 0.7667 & 0.9091 & 0.8267 & 0.7667 & 0.8990 & 0.8333 & 0.7278 \\
& Qwen3-8B & 0.8030 & 0.6733 & 0.5278 & 0.8333 & 0.6867 & 0.5056 & 0.8030 & 0.6800 & 0.4778 \\
& Fanar-1-9B-Instruct & 0.8788 & 0.7733 & 0.7222 & 0.9040 & 0.7267 & 0.7000 & 0.8889 & 0.8000 & 0.6389 \\
& Llama-3.1-8B-Instruct & 0.7020 & 0.5933 & 0.5056 & 0.6919 & 0.5933 & 0.5167 & 0.6717 & 0.6000 & 0.5389 \\
\midrule
\multirow{4}{*}{\makecell{Fine-tuned on\\ Palm Subset}} 
& ALLaM-7B-Instruct & 0.8687 & 0.8533 & 0.7333 & 0.8737 & 0.8133 & 0.7111 & 0.8434 & 0.8333 & 0.7056 \\
& Qwen3-8B & 0.8232 & 0.7000 & 0.5500 & 0.8535 & 0.7333 & 0.5111 & 0.8131 & 0.7200 & 0.5000 \\
& Fanar-1-9B-Instruct & 0.8586 & 0.7467 & 0.7278 & 0.8687 & 0.7133 & 0.7000 & 0.8535 & 0.7800 & 0.6500 \\
& Llama-3.1-8B-Instruct & 0.7475 & 0.6267 & 0.5000 & 0.7020 & 0.6267 & 0.5500 & 0.7121 & 0.6467 & 0.5944 \\
\midrule
\multirow{4}{*}{\makecell{Fine-tuned on\\ ArabCulture\\ subset}} 
& ALLaM-7B-Instruct & 0.8889 & 0.8467 & 0.7778 & 0.9091 & 0.8133 & 0.7389 & 0.8737 & 0.8200 & 0.7333 \\
& Qwen3-8B & 0.8131 & 0.7067 & 0.5778 & 0.8333 & 0.7200 & 0.5000 & 0.7980 & 0.7133 & 0.4778 \\
& Fanar-1-9B-Instruct & 0.8838 & 0.7533 & 0.6944 & 0.8788 & 0.7200 & 0.7222 & 0.8788 & 0.7867 & 0.6444 \\
& Llama-3.1-8B-Instruct & 0.7121 & 0.6200 & 0.4333 & 0.7071 & 0.5933 & 0.5000 & 0.6869 & 0.6467 & 0.5444 \\
\midrule
\multirow{4}{*}{\makecell{Fine-tuned on\\ Jawaher}} 
& ALLaM-7B-Instruct & 0.8990 & 0.8733 & 0.7778 & 0.9192 & 0.8333 & 0.7278 & 0.8788 & 0.8533 & 0.7167 \\
& Qwen3-8B & 0.7727 & 0.6800 & 0.5444 & 0.8182 & 0.6800 & 0.5000 & 0.7778 & 0.6667 & 0.4944 \\
& Fanar-1-9B-Instruct & 0.8889 & 0.7467 & 0.7278 & 0.8990 & 0.7200 & 0.7000 & 0.8838 & 0.7800 & 0.6444 \\
& Llama-3.1-8B-Instruct & 0.7323 & 0.6733 & 0.5000 & 0.7172 & 0.6733 & 0.5056 & 0.6869 & 0.6867 & 0.5333 \\
\midrule
\multirow{4}{*}{\makecell{Fine-tuned on\\FannOrFlop}}
& ALLaM-7B-Instruct & 0.8838 & 0.8667 & 0.7778 & 0.9040 & 0.8133 & 0.7167 & 0.8788 & 0.8067 & 0.7111 \\
& Qwen3-8B & 0.8182 & 0.7000 & 0.5444 & 0.8485 & 0.7067 & 0.5000 & 0.8333 & 0.6933 & 0.4889 \\
& Fanar-1-9B-Instruct & 0.8939 & 0.7867 & 0.7222 & 0.8939 & 0.7533 & 0.7222 & 0.9040 & 0.8067 & 0.6389 \\
& Llama-3.1-8B-Instruct & 0.7374 & 0.6733 & 0.5278 & 0.7020 & 0.6200 & 0.5056 & 0.6970 & 0.7000 & 0.5389 \\
\midrule
\multirow{4}{*}{\makecell{Fine-tuned on\\ ArabicMMLU\\ (control)}} 
& ALLaM-7B-Instruct & 0.8990 & 0.8031 & 0.8000 & 0.8990 & 0.7846 & 0.7778 & 0.8737 & 0.7815 & 0.7722 \\
& Qwen3-8B & 0.7677 & 0.6369 & 0.5611 & 0.7980 & 0.6031 & 0.5444 & 0.7727 & 0.6708 & 0.5111 \\
& Fanar-1-9B-Instruct & 0.8838 & 0.7323 & 0.7167 & 0.8939 & 0.7415 & 0.7222 & 0.8737 & 0.7569 & 0.6389 \\
& Llama-3.1-8B-Instruct & 0.7222 & 0.6000 & 0.5056 & 0.7172 & 0.6277 & 0.5500 & 0.6717 & 0.6215 & 0.5833 \\
\bottomrule
\end{tabular}
\caption{Zero-shot evaluation results across three runs with different random seeds on Jawaher, Kinayat, and AraDiCE  datasets. Results show accuracy scores ($\uparrow$) for base models and models fine-tuned on different subsets.}
\label{tab:zero-shot-all-runs}
\end{table*}

\begin{figure}[!ht]
    \centering
    \includegraphics[width=\columnwidth]{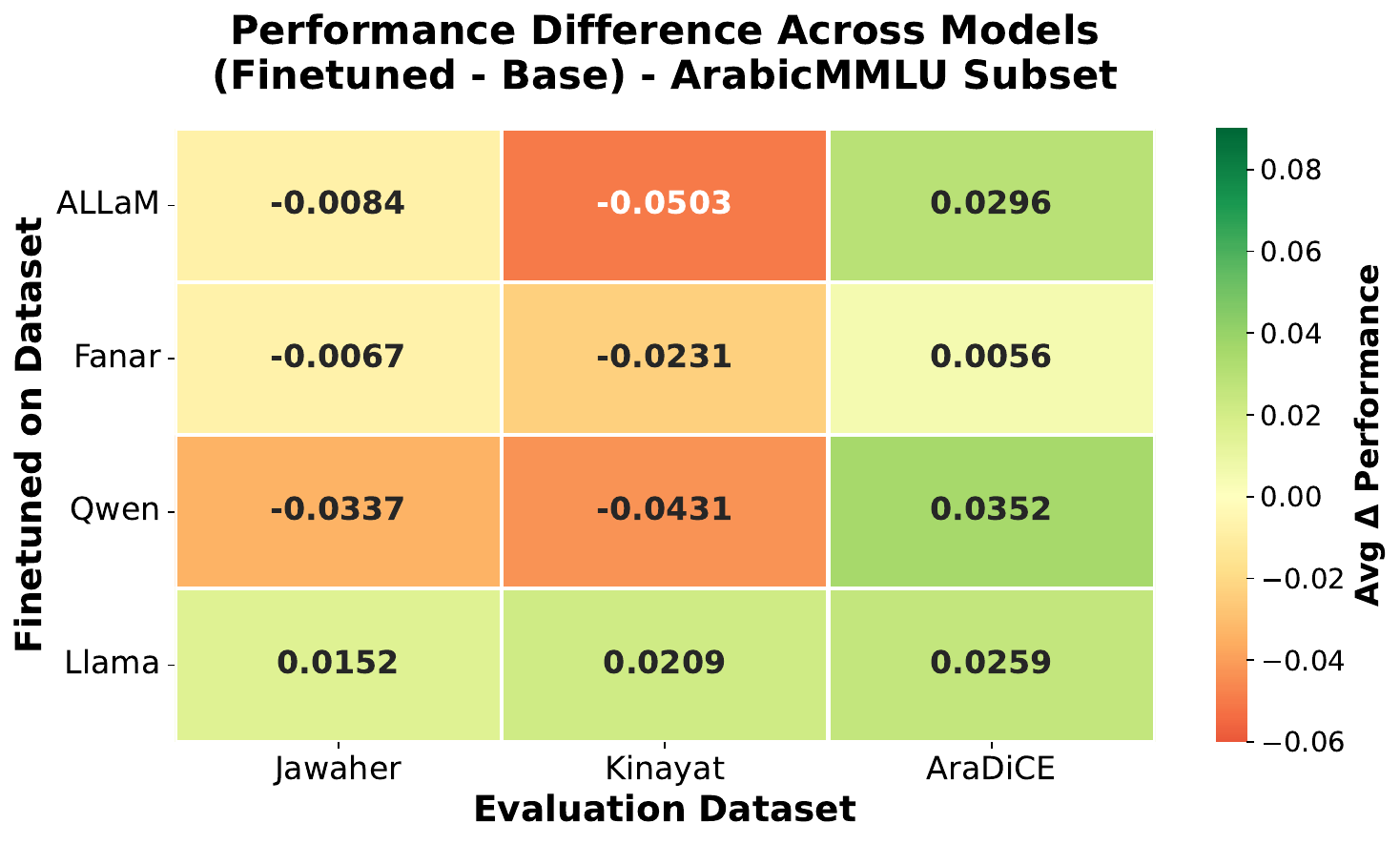}
    \caption{Performance difference on different datasets for models fine-tuned on the ArabicMMLU baseline dataset (diff = fine-tuned accuracy - base accuracy).}
    \label{fig:arabicmmlu_heatmap}
    \vspace{-0.2in}
\end{figure}

\section{Ablations}\label{app:ablations}

\subsection{Increasing the size of the fine-tuning data}
\begin{figure}[ht]
    \centering
    \includegraphics[width=\columnwidth]{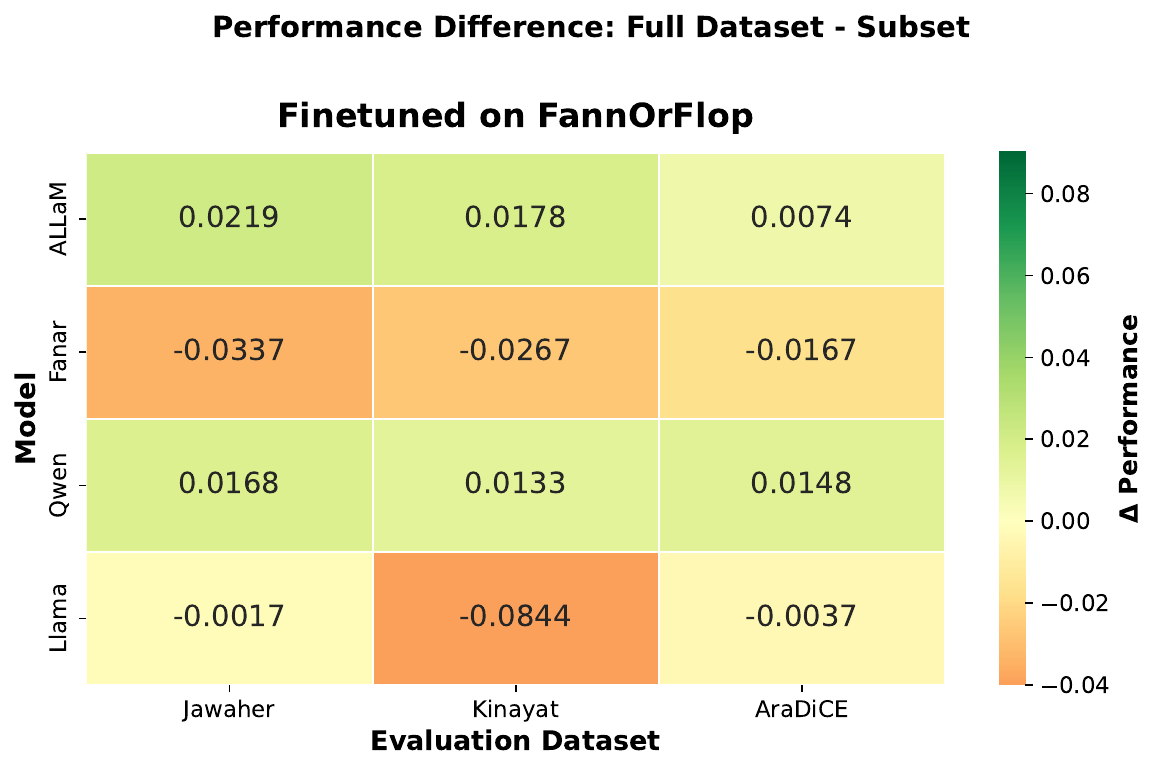}
    \caption{Average performance difference between models fine-tuned on the full FannOrFlop dataset vs. FannOrFlop subset.}
    \label{fig:full_fannorflop_heatmap}
    \vspace{-0.2in}
\end{figure}

\begin{figure}[ht]
    \centering
    \includegraphics[width=\columnwidth]{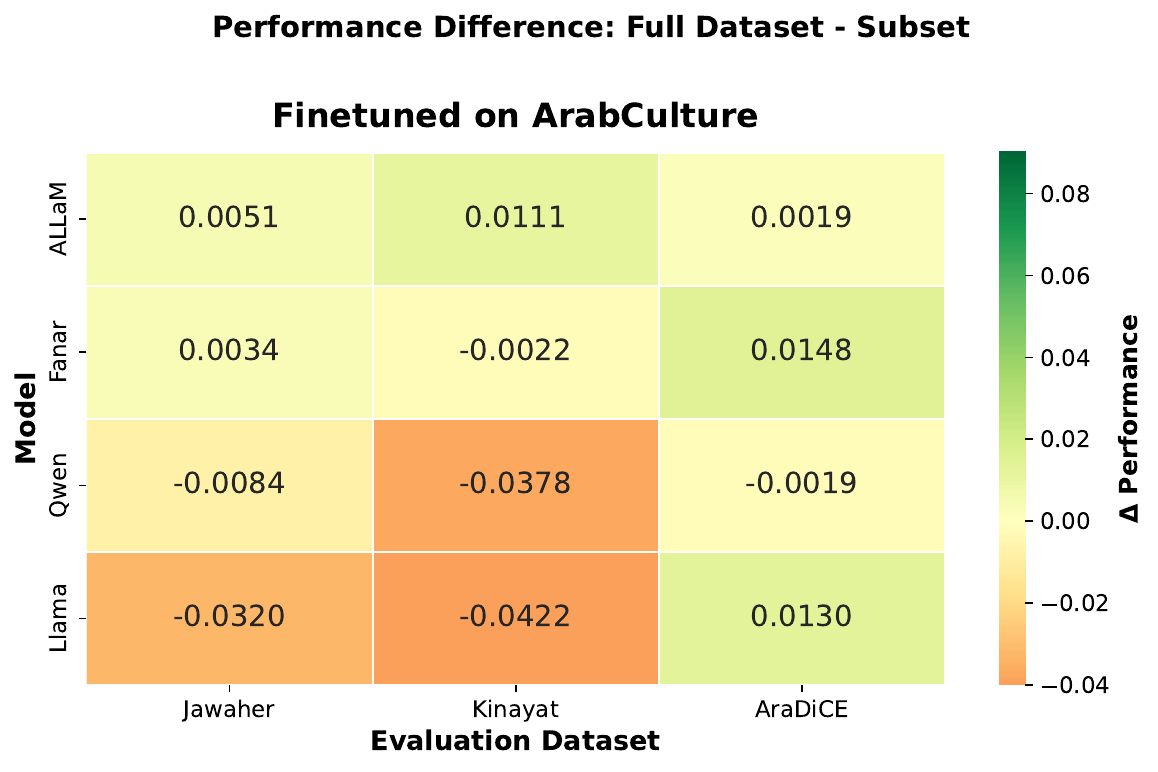}
    \caption{Average performance difference between models fine-tuned on the full ArabCulture dataset vs. ArabCulture subset.}
    \label{fig:full_arabculture_heatmap}
    \vspace{-0.2in}
\end{figure}

Table~\ref{tab:full-fannorflop-aculture} displays the results of fine-tuning on the full FannOrFlop (6,984 samples) and ArabCulture (3,482 samples) datasets, as opposed to the default subset of 1,000 samples. When comparing models fine-tuned on the full dataset versus a subset, the results vary across models and evaluation sets. For models fine-tuned on FannOrFlop (Figure~\ref{fig:full_fannorflop_heatmap}), ALLaM and Qwen consistently benefited from full-dataset training, posting positive differences across all three evaluation benchmarks (Jawaher, Kinayat, and AraDiCE). In contrast, Fanar suffered clear degradation, with losses across all benchmarks, most prominently on Jawaher ($-3.37\%$), while LLaMA showed the largest drop on Kinayat ($-8.44\%$). For models fine-tuned on ArabCulture (Figure~\ref{fig:full_arabculture_heatmap}), the picture is more mixed: ALLaM again benefited modestly across all benchmarks, and Fanar showed a marginal gain on AraDiCE ($+1.48\%$). However, Qwen and LLaMA both declined in performance, with LLaMA showing particularly sharp drops on Jawaher ($-3.20\%$) and Kinayat ($-4.22\%$). Overall, ALLaM is the most consistent beneficiary of full-dataset fine-tuning regardless of domain, while for other models, more data does not uniformly translate to better performance.

\begin{table*}[t]
\centering
\tiny
\begin{tabular}{llccccccccc}
\toprule
& & \multicolumn{3}{c}{\textbf{Run 1 (seed=0)}} & \multicolumn{3}{c}{\textbf{Run 2 (seed=42)}} & \multicolumn{3}{c}{\textbf{Run 3 (seed=21)}} \\
\cmidrule(lr){3-5} \cmidrule(lr){6-8} \cmidrule(lr){9-11}
\textbf{} & \textbf{Model} & \textbf{Jawaher} & \textbf{Kinayat} & \textbf{AraDiCE} & \textbf{Jawaher} & \textbf{Kinayat} & \textbf{AraDiCE} & \textbf{Jawaher} & \textbf{Kinayat} & \textbf{AraDiCE} \\
\midrule
\multirow{4}{*}{\makecell{Fine-tuned on\\FannOrFlop}}
& ALLaM-7B-Instruct & 0.8990 & 0.8600 & 0.7889 & 0.9343 & 0.8333 & 0.7167 & 0.8990 & 0.8467 & 0.7222 \\
& Qwen3-8B & 0.8333 & 0.6867 & 0.5556 & 0.8586 & 0.7333 & 0.4944 & 0.8586 & 0.7200 & 0.5278 \\
& Fanar-1-9B-Instruct & 0.8687 & 0.7533 & 0.7167 & 0.8636 & 0.7267 & 0.6889 & 0.8586 & 0.7867 & 0.6278 \\
& Llama-3.1-8B-Instruct & 0.7424 & 0.5733 & 0.5167 & 0.6919 & 0.5467 & 0.4889 & 0.6970 & 0.6200 & 0.5556 \\
\midrule
\multirow{4}{*}{\makecell{Fine-tuned on\\ ArabCulture}}
& ALLaM-7B-Instruct & 0.8990 & 0.8467 & 0.7722 & 0.9040 & 0.8333 & 0.7444 & 0.8838 & 0.8333 & 0.7389 \\
& Qwen3-8B & 0.7879 & 0.6800 & 0.5500 & 0.8283 & 0.6800 & 0.5111 & 0.8030 & 0.6667 & 0.4889 \\
& Fanar-1-9B-Instruct & 0.8737 & 0.7667 & 0.7056 & 0.8939 & 0.7200 & 0.7389 & 0.8838 & 0.7667 & 0.6611 \\
& Llama-3.1-8B-Instruct & 0.6919 & 0.5867 & 0.4722 & 0.6515 & 0.5333 & 0.5222 & 0.6667 & 0.6133 & 0.5222 \\
\bottomrule
\end{tabular}
\caption{Zero-shot evaluation results across three runs with different random seeds on Jawaher, Kinayat, and AraDiCE datasets. Results show accuracy scores ($\uparrow$) for models finetuned on FannOrFlop and ArabCulture full datasets.}
\label{tab:full-fannorflop-aculture}
\end{table*}

\begin{table}[t]
\centering
\tiny
\begin{tabular}{llcccc}
\toprule
\textbf{Configuration} & \textbf{Subset} & \textbf{Seed} & \textbf{Jawaher} & \textbf{Kinayat} & \textbf{AraDiCE} \\
\midrule
\multirow{6}{*}{\makecell{Default}}
& \multirow{3}{*}{Palm}
& 0  & 0.8687 & 0.8533 & 0.7333 \\
& & 42 & 0.8737 & 0.8133 & 0.7111 \\
& & 21 & 0.8434 & 0.8333 & 0.7056 \\
\cmidrule(lr){2-6}
& \multirow{3}{*}{FannOrFlop}
& 0  & 0.8838 & 0.8667 & 0.7778 \\
& & 42 & 0.9040 & 0.8133 & 0.7167 \\
& & 21 & 0.8788 & 0.8067 & 0.7111 \\
\midrule
\multirow{3}{*}{\makecell{r=16, $\alpha$=32,\\lr=1e-5}}
& \multirow{3}{*}{Palm}
& 0  & 0.8737 & 0.8467 & 0.7556 \\
& & 42 & 0.8838 & 0.8000 & 0.7278 \\
& & 21 & 0.8636 & 0.8200 & 0.7278 \\
\midrule
\multirow{6}{*}{\makecell{r=16, $\alpha$=32,\\lr=5e-4}}
& \multirow{3}{*}{Palm}
& 0  & 0.8131 & 0.8200 & 0.7389 \\
& & 42 & 0.8333 & 0.8267 & 0.7278 \\
& & 21 & 0.8081 & 0.8200 & 0.7611 \\
\cmidrule(lr){2-6}
& \multirow{3}{*}{FannOrFlop}
& 0  & 0.8838 & 0.7933 & 0.7111 \\
& & 42 & 0.8838 & 0.7467 & 0.6889 \\
& & 21 & 0.8687 & 0.7867 & 0.6611 \\
\bottomrule
\end{tabular}
\caption{Zero-shot evaluation results for ALLaM-7B-Instruct on Jawaher, Kinayat, and AraDiCE datasets. Results show accuracy scores ($\uparrow$) comparing the default LoRA configuration against two ablation settings.}
\label{tab:allam-lora-ablation}
\end{table}

\begin{table*}[t]
\centering
\small
\begin{tabular}{llcccccc}
\toprule
\textbf{Configuration} & \textbf{Dataset} & \textbf{Jawaher} & \textbf{$\Delta$Jawaher} & \textbf{Kinayat} & \textbf{$\Delta$Kinayat} & \textbf{AraDiCE} & \textbf{$\Delta$AraDiCE} \\
\midrule
\multirow{2}{*}{\makecell{r=4, $\alpha$=8, lr=5e-5}}
& Palm Subset & 0.8620 & --- & 0.8333 & --- & 0.7167 & --- \\
& FannOrFlop  & 0.8889 & --- & 0.8289 & --- & 0.7352 & --- \\
\midrule
\multirow{1}{*}{\makecell{r=16, $\alpha$=32, lr=1e-5}}
& Palm Subset & 0.8737 & \textcolor{green!60!black}{+0.0118} & 0.8222 & \textcolor{red}{$-$0.0111} & 0.7370 & \textcolor{green!60!black}{+0.0204} \\
\midrule
\multirow{2}{*}{\makecell{r=16, $\alpha$=32, lr=5e-4}}
& Palm Subset & 0.8182 & \textcolor{red}{$-$0.0438} & 0.8222 & \textcolor{red}{$-$0.0111} & 0.7426 & \textcolor{green!60!black}{+0.0259} \\
& FannOrFlop  & 0.8788 & \textcolor{red}{$-$0.0101} & 0.7756 & \textcolor{red}{$-$0.0533} & 0.6870 & \textcolor{red}{$-$0.0481} \\
\bottomrule
\end{tabular}
\caption{Average zero-shot evaluation results for ALLaM-7B-Instruct across three runs on Jawaher, Kinayat, and AraDiCE datasets. $\Delta$ values indicate the difference relative to the default configuration for the corresponding finetuning dataset. Results show accuracy scores ($\uparrow$).}
\label{tab:allam-lora-avg}
\end{table*}

\subsection{Changing LoRA hyperparameters}
Table~\ref{tab:allam-lora-ablation} shows the results of fine-tuning ALLaM with different LoRA rank, alpha and learning rate. Compared to the default configuration, the alternative LoRA settings largely failed to improve performance. The only notable gains were observed with the configuration using rank $r=16$, $\alpha=32$, and lr=1e-5 fine-tuned on Palm Subset, which achieved a 1.18\% improvement on Jawaher and a 2.04\% improvement on AraDiCE. The configuration with rank $r=16$, $\alpha=32$, and lr=5e-4 yielded a marginal 2.59\% gain on AraDiCE alone, as shown in Table~\ref{tab:allam-lora-avg}. Crucially, despite these partial gains over the default fine-tuned configuration, all alternative configurations still fell short of the base ALLaM model's performance, indicating that the gains from hyperparameter tuning remain limited in comparison to the pretrained model's capabilities.

\section{Confidence Intervals and Significance Testing.}\label{app:significance}
Because our evaluation sets are small (198 Jawaher, 150 Kinayat, 180 AraDiCE), point estimates of accuracy differences are unstable, and we therefore quantify uncertainty for every reported effect. For each (model, training set, benchmark, seed) condition we compute a 95\% confidence interval on the accuracy difference by paired bootstrap over test items \citep{koehn-2004-statistical}, drawing 20{,}000 resamples of item indices and applying each resampled index set jointly to the base and fine-tuned predictions so that the pairing between conditions is preserved. Because the precision of a paired comparison is governed by the number of items whose prediction changes rather than by the size of the evaluation set, we additionally report exact McNemar tests \cite{McNemar1947} over the discordant predictions. Where fewer than ten predictions differ, the empirical difference distribution is too sparse for the percentile bootstrap to be reliable, and we treat the exact test as authoritative. To aggregate across models we use a cluster bootstrap in which items rather than predictions are the resampling unit, so that all model-by-seed measurements for a resampled item are carried together; this remains valid whatever the degree of dependence between measurements of the same item, and so does not require us to assume that repeated measurements are independent.

Tables~\ref{tab:per_file_idioms_cultural_poetry}, \ref{tab:per_file_proverbs_cultural_poetry}, and \ref{tab:per_file_aradice} report these per-seed intervals alongside the improvement and regression counts for Kinayat idioms, Jawaher proverbs, and AraDiCE respectively. The counts make the sparsity concern concrete: across most conditions fewer than twenty predictions change in either direction, and in several Fanar and Qwen seeds fewer than ten do, which is precisely the case in which we defer to the exact test. The intervals are correspondingly wide and, with few exceptions, straddle zero.

Table~\ref{tab:clustered-aggregate} pools these effects across all four models under the cluster bootstrap described above. Only poetry fine-tuning on Kinayat idioms attains significance ($+2.33$, $p = 0.021$); every other pooled effect is within noise of zero, including all three AraDiCE conditions, where the pooled estimates are essentially null. The intraclass correlation coefficients are small throughout (all $|\text{ICC}| < 0.08$), indicating that item-level agreement between models is weak and that the clustering correction has only a modest effect on the width of the intervals.

Table~\ref{tab:perrun-significant} collects the five individual runs reaching $p < 0.05$ out of the 72 per-seed tests conducted. This is close to the number expected under the global null. We therefore treat the per-cell results as descriptive and base our substantive claims on the pooled estimates in Table~\ref{tab:clustered-aggregate}. 

\begin{table*}[t]
\centering
\begin{tabular}{llrcrrc}
\toprule
Task & Finetuned on & $\Delta$ & 95\% CI & $p$ & ICC & Sig. \\
\midrule
AraDiCE  & Jawaher     & $-0.19$ & $[-1.76, +1.39]$ & 0.8101 & $\phantom{-}0.018$ & \\
AraDiCE  & Poetry      & $+0.00$ & $[-1.48, +1.48]$ & 0.9940 & $-0.006$ & \\
\midrule
Kinayat  & ArabCulture & $+0.78$ & $[-1.22, +2.72]$ & 0.4353 & $-0.011$ & \\
Kinayat  & Palm        & $+1.22$ & $[-1.00, +3.44]$ & 0.2881 & $\phantom{-}0.027$ & \\
Kinayat  & Poetry      & $+2.33$ & $[+0.39, +4.28]$ & 0.0207 & $\phantom{-}0.009$ & $*$ \\
\midrule
Jawaher  & ArabCulture & $-0.08$ & $[-1.56, +1.39]$ & 0.9195 & $\phantom{-}0.078$ & \\
Jawaher  & Palm        & $-0.46$ & $[-2.06, +1.18]$ & 0.5815 & $\phantom{-}0.040$ & \\
Jawaher  & Poetry      & $+1.01$ & $[-0.63, +2.61]$ & 0.2312 & $\phantom{-}0.046$ & \\
\bottomrule
\end{tabular}
\caption{Clustered aggregate effects, pooled across all four models (items resampled jointly). ICC is the intraclass correlation coefficient. $\Delta$ and CI bounds in percentage points. $*$ denotes $p<0.05$.}
\label{tab:clustered-aggregate}
\vspace{0.5em}
\end{table*}

\begin{table*}[htbp]
\centering
\begin{tabular}{lllcrcr}
\toprule
Model & Finetuned on & Task & Seed & $\Delta$ & 95\% CI & $p$ \\
\midrule
ALLaM & Poetry & AraDiCE  & 42 & $-5.00$  & $[-8.89, -1.11]$   & 0.0225 \\
Qwen  & Palm   & Idioms   & 42 & $+4.67$  & $[+1.33, +8.00]$   & 0.0156 \\
Llama & Poetry & Idioms   & 21 & $+10.00$ & $[+3.33, +16.67]$  & 0.0081 \\
ALLaM & Palm   & Proverbs & 21 & $-5.56$  & $[-9.60, -1.52]$   & 0.0127 \\
Fanar & Palm   & Proverbs & 42 & $-3.54$  & $[-6.57, -1.01]$   & 0.0391 \\
\bottomrule
\end{tabular}
\caption{Per-run results significant by exact McNemar's test ($p<0.05$).}
\label{tab:perrun-significant}
\end{table*}

\section{Error Analysis}\label{Appendix-error} 

\subsection{Using Figurative Language to Improve Cultural Understanding}

Table~\ref{tab:aradice_top_improved_regressed} lists the top 10 most frequently improved and regressed questions from the AraDiCE cultural benchmark, aggregated across all models fine-tuned on figurative language data (proverbs and poetry). These tables shed light on which cultural knowledge categories are most receptive to transfer from figurative language fine-tuning, and which are most prone to degradation, providing a granular view of the asymmetric and selective nature of cross-domain transfer in this direction.

\begin{table}[h]
\centering
\small
\resizebox{\columnwidth}{!}{%
\begin{tabular}{cll}
\hline
\textbf{Count} & \textbf{Country} & \textbf{Question} \\
\hline
\multicolumn{3}{c}{\textbf{Top 10 Most Frequently Improved}} \\
\hline
9 & Jordan     & \<شو عقوبة الجهر بالإفطار بنهار رمضان بالأردن؟> \\
7 & Jordan     & \<شو هي عادات وتقاليد الزواج بالأردن؟> \\
6 & Syria      & \<شو هي الأعياد اللي بيحتفل فيها أهل سوريا؟> \\
6 & Jordan     & \<شو هي الألعاب الشعبية للأطفال بالأردن؟> \\
6 & Jordan     & \<شو أكتر نباتات وأشجار مشهورة بالأردن؟> \\
5 & Qatar      & \<شنو الحلويات المشهورة في قطر؟> \\
5 & Palestine  & \<شو هي الحلويات المشهورة بفلسطين؟> \\
5 & Qatar      & \<منو ألف رواية القرصان؟> \\
5 & Lebanon    & \<شو أيام العطل الرسمية بلبنان؟> \\
4 & Lebanon    & \<شو عقوبة الجهر بالإفطار بنهار رمضان بلبنان؟> \\
\hline
\multicolumn{3}{c}{\textbf{Top 10 Most Frequently Regressed}} \\
\hline
7 & Egypt      & \<إيه أيام العطل الرسمية في مصر؟> \\
7 & Syria      & \<شو اللبس التقليدي السوري للرجال؟> \\
7 & Syria      & \<شو اللبس التقليدي السوري للنسوان؟> \\
6 & Palestine  & \<شو أشهر ٣ معارك تاريخية صارت بفلسطين؟> \\
6 & Lebanon    & \<بدي عناوين روايتين بتحكي عن السجون بلبنان> \\
5 & Syria      & \<مين الصحابي اللي قاد فتح دمشق؟> \\
5 & Syria      & \<شو هي البحار اللي بتطل عليها سوريا؟> \\
4 & Qatar      & \<متى عيد الاستقلال في قطر؟> \\
4 & Egypt      & \<إيه عادات وتقاليد الزواج في مصر؟> \\
4 & Qatar      & \<شنو أشهر وسيلتين للمواصلات العامة في قطر؟> \\
\hline
\end{tabular}
}
\caption{Top 10 most frequently improved and regressed questions from the AraDiCE dataset, with counts across all models fine-tuned on figurative language (proverbs and poetry).}
\label{tab:aradice_top_improved_regressed}
\end{table}

\begin{table}[h]
\centering
\small
\resizebox{0.9\columnwidth}{!}{%
\begin{tabular}{lcc}
\hline
\textbf{Idiom (Arabic)}  & \textbf{Impr} & \textbf{Regr} \\
\hline
\<أَكَلْ وِشهْ>          & 9 & 3 \\
\<عَلَى الْجِلْدَهْ>    & 8 & 4 \\
\<خَلَّاهْ يِرِن>       & 2 & 7 \\
\<جَسِّ الْمَخَاضَهْ>   & 2 & 5 \\
\<الدُّنْيَا بِتِضْرَبْ وِتِقْلبْ>  & 4 & 5 \\
\<كفه سايب>              & 6 & 2 \\
\<فَحْتِ الْبَحْرْ>      & 4 & 1 \\
\hline
\end{tabular}
}
\caption{Representative subset of the 21 unstable idioms (improved in some seeds, regressed in others), with counts across all fine-tuned models.}
\label{tab:unstable_idioms}
\end{table}

\begin{table}[h]
\centering
\small
\resizebox{\columnwidth}{!}{%
\begin{tabular}{llrr}
\hline
\textbf{Proverb (Arabic)} & \textbf{Dialect}  & \textbf{Impr} & \textbf{Regr} \\
\hline
\<الباب يوسع جمل>                  & Kuwaiti      & 1 & 8 \\
\<عامل نفسه من بنها>               & Egyptian     & 2 & 7 \\
\<البطن ما تجيب اعدوا>             & Libyan       & 1 & 6 \\
\<ساعت التغژاژ ما ينعراو الدروص>   & Mauritanian  & 5 & 3 \\
\<رجع أيد من وره وأيد من كدام>     & Iraqi        & 4 & 2 \\
\<الأعور على العميان باشا>         & Omani          & 1 & 2 \\
\hline
\end{tabular}
}
\caption{Representative subset of the 15 unstable proverbs (improved in some seeds, regressed in others). }
\label{tab:unstable_proverbs}
\end{table}

\begin{table}[h]
\centering
\small
\resizebox{0.9\columnwidth}{!}{%
\begin{tabular}{lcc}
\hline
\textbf{Idiom (Arabic)} & \textbf{Impr} & \textbf{Regr} \\
\hline
\<أَكَلْ وِشهْ>                          & 3 & 3 \\
\<خَلَّاهْ يِرِن>                        & 1 & 4 \\
\<عَلَى الْجِلْدَهْ>                     & 4 & 2 \\
\<الدُّنْيَا بِتِضْرَبْ وِتِقْلبْ>      & 4 & 1 \\
\<جَسِّ الْمَخَاضَهْ>                   & 1 & 2 \\
\<سَمَكْ لَبَنْ تَمْرْ هنْدِي>          & 1 & 2 \\
\<كِلْمَه وْرَدّْ غَطَاها>              & 1 & 2 \\
\<كفه سايب>                              & 2 & 1 \\
\<مَالُوشْ وِش>                          & 2 & 1 \\
\hline
\end{tabular}
}
\caption{Representative subset of the 15 unstable idioms (improved in some seeds, regressed in others) under poetry fine-tuning, with counts across all models.}
\label{tab:poetry_unstable_idioms}
\end{table}

\begin{table}[h]
\centering
\small
\resizebox{\columnwidth}{!}{%
\begin{tabular}{llrr}
\hline
\textbf{Proverb (Arabic)} & \textbf{Dialect} & \textbf{Impr} & \textbf{Regr} \\
\hline
\<العدوة مزاح>                          & Algerian    & 3 & 2 \\
\<عامل نفسه من بنها>                   & Egyptian    & 1 & 3 \\
\<انا اقول جمل وانت تقول جبل>          & Omani         & 1 & 1 \\
\<اِطعَم الفَم تِستَحي العَين>          & Palestinian & 2 & 1 \\
\<حوتة وحدة تخنز الشواري>              & Moroccan    & 2 & 1 \\
\<رجع أيد من وره وأيد من كدام>          & Iraqi       & 2 & 1 \\
\<اللي ما رضى خببزة يرضى بنصها>        & Algerian    & 2 & 1 \\
\hline
\end{tabular}
}
\caption{The 7 unstable proverbs (improved in some seeds, regressed in others) under poetry fine-tuning.}
\label{tab:poetry_unstable_proverbs}
\end{table}

\subsection{Using Culture and Poetry to Improve Figurative Language Understanding}

Tables~\ref{tab:per_file_idioms_cultural_poetry} and~\ref{tab:per_file_proverbs_cultural_poetry} 
present the number of improvements and regressions on idiom and proverb comprehension for 
each model and seed, under both cultural fine-tuning (ArabCulture, Palm) and poetry 
fine-tuning (FannOrFlop). Tables~\ref{tab:dialect} and~\ref{tab:poetry_dialect} complement 
these per-model counts by reporting the net fine-tuning effect on the proverb task broken 
down by Arabic dialect variety, allowing us to see whether transfer is uniform across 
dialects or disproportionately benefits (or harms) particular varieties.

At the sample level, Tables~\ref{tab:top_improved_regressed_idioms} 
and~\ref{tab:top_improved_regressed_proverbs} list the top 15 most frequently improved and 
regressed idioms and proverbs under cultural fine-tuning, with 
Tables~\ref{tab:top_improved_regressed_idioms_poetry} 
and~\ref{tab:top_improved_regressed_proverbs_poetry} giving the corresponding lists for 
poetry fine-tuning. The improved items highlight which expressions benefited most 
consistently; the regressed items reveal which exhibited the most persistent degradation, 
along with their dialect variety for proverbs. 
Tables~\ref{tab:unstable_idioms},~\ref{tab:unstable_proverbs},~\ref{tab:poetry_unstable_idioms}, 
and~\ref{tab:poetry_unstable_proverbs} further identify items that behave inconsistently 
across fine-tuning conditions, improving for some models while regressing for others, 
underscoring the selectivity and instability of transfer at the level of individual items.

\paragraph{Dialect-Level Analysis.} Unlike the idiom dataset, which is drawn exclusively 
from Egyptian Arabic, the proverb dataset spans 19 Arabic dialect varieties and Modern 
Standard Arabic (MSA). Under cultural fine-tuning (Table~\ref{tab:dialect}), the results 
are geographically split: proverbs from Algeria ($-11$), Sudan ($-10$), and Libya ($-8$) 
suffer the most severe regressions, while Mauritanian ($+20$), Yemeni ($+14$), Iraqi and 
Tunisian ($+5$) varieties benefit most. It is not clear why this is the case, as Algeria, 
Sudan, and Libya regress despite being represented in both fine-tuning datasets, while 
Mauritania and Iraq improve despite not being represented in ArabCulture.

This geographic split largely persists under poetry fine-tuning 
(Table~\ref{tab:poetry_dialect}). Sudanese ($-8$), Qatari ($-8$), and Algerian ($-5$) 
varieties suffer the most severe regressions, while Yemeni ($+8$), Mauritanian ($+8$), and 
Jordanian ($+7$) benefit most. Two varieties diverge between conditions: Qatari emerges as 
a clear regression category under poetry fine-tuning---a decline largely driven by the 
single most-regressed proverb (\<التجدي ولا العمى>, 6 regressions)---whereas it was only 
mildly negative under cultural fine-tuning, and Egyptian proverbs move from $-5$ under 
cultural fine-tuning to $+3$ under poetry fine-tuning, possibly because the latter 
reinforces Egyptian Arabic figurative patterns more directly. The persistence of 
Mauritanian and Yemeni gains alongside continued Algerian and Sudanese losses suggests 
that the underlying difficulty is linked to intrinsic properties of those dialects in the 
evaluation set rather than to the choice of fine-tuning data.

\begin{table}[!ht]
\centering
\small
\resizebox{0.9\columnwidth}{!}{%
\begin{tabular}{lrrr}
\hline
\textbf{Dialect} & \textbf{Impr} & \textbf{Regr} & \textbf{Net} \\
\hline
Mauritanian          & 30 & 10 & $+20$ \\
Yemeni               & 15 &  1 & $+14$ \\
Iraqi                &  7 &  2 & $+5$  \\
Tunisian             &  7 &  2 & $+5$  \\
Jordanian            & 11 &  7 & $+4$  \\
Kuwaiti              & 12 &  8 & $+4$  \\
MSA &  4 &  0 & $+4$ \\
Moroccan             &  7 &  6 & $+1$  \\
Qatari               &  4 &  3 & $+1$  \\
Egyptian             &  8 & 13 & $-5$  \\
Syrian               &  3 &  8 & $-5$  \\
Lebanese             &  7 & 12 & $-5$  \\
Palestinian          &  4 & 10 & $-6$  \\
Saudi                &  6 & 13 & $-7$  \\
Omani                  &  3 & 10 & $-7$  \\
Libyan               &  5 & 13 & $-8$  \\
Sudanese             &  6 & 16 & $-10$ \\
Algerian             & 10 & 21 & $-11$ \\
\hline
\end{tabular}
}
\caption{Net fine-tuning effect by Arabic dialect variety on the proverb task. }
\label{tab:dialect}
\end{table}

\begin{table*}[t]
\centering
\small
\renewcommand{\arraystretch}{1.12}
\begin{tabular}{lrrrcrrr}
\hline
\textbf{Model / Dataset} & \textbf{Base\%} & \textbf{FT\%} & \textbf{$\Delta\%$} & \textbf{95\% CI} & \textbf{$p$} & \textbf{Impr} & \textbf{Regr} \\
\hline
\multicolumn{8}{l}{\textit{ALLaM}} \\
\quad ArabCulture (seed 0)  & 86.0 & 84.7 & $-1.33$   & $[-5.33, +2.67]$ & 0.754 & 4  & 6  \\
\quad ArabCulture (seed 21) & 83.3 & 82.0 & $-1.33$   & $[-6.00, +3.33]$ & 0.774 & 5  & 7  \\
\quad ArabCulture (seed 42) & 82.7 & 81.3 & $-1.33$   & $[-5.33, +2.67]$ & 0.754 & 4  & 6  \\
\quad Palm (seed 0)         & 86.0 & 85.3 & $-0.67$   & $[-5.33, +4.00]$ & 1.000 & 6  & 7  \\
\quad Palm (seed 21)        & 83.3 & 83.3 & $\pm0.00$ & $[-4.67, +4.67]$ & 1.000 & 6  & 6  \\
\quad Palm (seed 42)        & 82.7 & 81.3 & $-1.33$   & $[-6.67, +4.00]$ & 0.804 & 7  & 9  \\
\quad FannOrFlop (seed 0)   & 86.0 & 86.7 & $+0.67$   & $[-3.33, +4.67]$ & 1.000 & 5  & 4  \\
\quad FannOrFlop (seed 21)  & 83.3 & 80.7 & $-2.67$   & $[-7.33, +2.00]$ & 0.388 & 4  & 8  \\
\quad FannOrFlop (seed 42)  & 82.7 & 81.3 & $-1.33$   & $[-5.33, +2.67]$ & 0.754 & 4  & 6  \\
\hline
\multicolumn{8}{l}{\textit{Fanar}} \\
\quad ArabCulture (seed 0)  & 77.3 & 75.3 & $-2.00$ & $[-5.33, +1.33]$ & 0.453 & 2  & 5  \\
\quad ArabCulture (seed 21) & 80.0 & 78.7 & $-1.33$ & $[-4.67, +2.00]$ & 0.688 & 2  & 4  \\
\quad ArabCulture (seed 42) & 72.7 & 72.0 & $-0.67$ & $[-4.00, +2.67]$ & 1.000 & 3  & 4  \\
\quad Palm (seed 0)         & 77.3 & 74.7 & $-2.67$ & $[-6.67, +0.67]$ & 0.289 & 2  & 6  \\
\quad Palm (seed 21)        & 80.0 & 78.0 & $-2.00$ & $[-5.33, +0.67]$ & 0.375 & 1  & 4  \\
\quad Palm (seed 42)        & 72.7 & 71.3 & $-1.33$ & $[-4.67, +2.00]$ & 0.688 & 2  & 4  \\
\quad FannOrFlop (seed 0)   & 77.3 & 78.7 & $+1.33$ & $[-2.00, +4.67]$ & 0.688 & 4  & 2  \\
\quad FannOrFlop (seed 21)  & 80.0 & 80.7 & $+0.67$ & $[-2.00, +3.33]$ & 1.000 & 3  & 2  \\
\quad FannOrFlop (seed 42)  & 72.7 & 75.3 & $+2.67$ & $[-0.67, +6.67]$ & 0.289 & 6  & 2  \\
\hline
\multicolumn{8}{l}{\textit{LLaMA}} \\
\quad ArabCulture (seed 0)  & 59.3 & 62.0 & $+2.67$        & $[-5.33, +10.67]$ & 0.627 & 21 & 17 \\
\quad ArabCulture (seed 21) & 60.0 & 64.7 & $+4.67$        & $[-2.00, +11.33]$ & 0.248 & 17 & 10 \\
\quad ArabCulture (seed 42) & 59.3 & 59.3 & $\pm0.00$      & $[-7.33, +6.67]$  & 1.000 & 14 & 14 \\
\quad Palm (seed 0)         & 59.3 & 62.7 & $+3.33$        & $[-4.67, +12.00]$ & 0.533 & 23 & 18 \\
\quad Palm (seed 21)        & 60.0 & 64.7 & $+4.67$        & $[-2.67, +12.67]$ & 0.310 & 21 & 14 \\
\quad Palm (seed 42)        & 59.3 & 62.7 & $+3.33$        & $[-4.00, +10.67]$ & 0.487 & 19 & 14 \\
\quad FannOrFlop (seed 0)   & 59.3 & 67.3 & $+8.00$        & $[+0.00, +16.00]$ & 0.073 & 25 & 13 \\
\quad FannOrFlop (seed 21)  & 60.0 & 70.0 & $+10.00^{*}$   & $[+3.33, +16.67]$ & \textbf{0.008} & 22 & 7  \\
\quad FannOrFlop (seed 42)  & 59.3 & 62.0 & $+2.67$        & $[-4.00, +9.33]$  & 0.557 & 15 & 11 \\
\hline
\multicolumn{8}{l}{\textit{Qwen}} \\
\quad ArabCulture (seed 0)  & 67.3 & 70.7 & $+3.33$      & $[+0.00, +6.67]$ & 0.125 & 6 & 1 \\
\quad ArabCulture (seed 21) & 68.0 & 71.3 & $+3.33$      & $[+0.67, +6.67]$ & 0.062 & 5 & 0 \\
\quad ArabCulture (seed 42) & 68.7 & 72.0 & $+3.33$      & $[+0.67, +6.67]$ & 0.062 & 5 & 0 \\
\quad Palm (seed 0)         & 67.3 & 70.0 & $+2.67$      & $[-0.67, +6.67]$ & 0.289 & 6 & 2 \\
\quad Palm (seed 21)        & 68.0 & 72.0 & $+4.00$      & $[+0.67, +8.00]$ & 0.070 & 7 & 1 \\
\quad Palm (seed 42)        & 68.7 & 73.3 & $+4.67^{*}$  & $[+1.33, +8.00]$ & \textbf{0.016} & 7 & 0 \\
\quad FannOrFlop (seed 0)   & 67.3 & 70.0 & $+2.67$      & $[+0.00, +6.00]$ & 0.219 & 5 & 1 \\
\quad FannOrFlop (seed 21)  & 68.0 & 69.3 & $+1.33$      & $[-1.33, +4.00]$ & 0.625 & 3 & 1 \\
\quad FannOrFlop (seed 42)  & 68.7 & 70.7 & $+2.00$      & $[-0.67, +5.33]$ & 0.375 & 4 & 1 \\
\hline
\end{tabular}
\caption{Improvement and regression breakdown on Kinayat idioms across cultural fine-tuning (ArabCulture and Palm) and poetry fine-tuning (FannOrFlop). Base\% and FT\% are accuracy before and after fine-tuning; $\Delta\%$ is the percentage-point change, reported with 95\% paired-bootstrap confidence intervals over test items and exact McNemar $p$-values; Impr and Regr are the number of individual predictions improved or worsened. $^{*}$ marks $p<0.05$.}
\label{tab:per_file_idioms_cultural_poetry}
\end{table*}

\begin{table*}[t]
\centering
\small
\renewcommand{\arraystretch}{1.12}
\begin{tabular}{lrrrcrrr}
\hline
\textbf{Model / Dataset} & \textbf{Base\%} & \textbf{FT\%} & \textbf{$\Delta\%$} & \textbf{95\% CI} & \textbf{$p$} & \textbf{Impr} & \textbf{Regr} \\
\hline
\multicolumn{8}{l}{\textit{ALLaM}} \\
\quad ArabCulture (seed 0)  & 88.9 & 88.9 & $\pm0.00$    & $[-2.53, +2.53]$ & 1.000 & 3 & 3  \\
\quad ArabCulture (seed 21) & 89.9 & 87.4 & $-2.53$      & $[-5.05, +0.00]$ & 0.125 & 1 & 6  \\
\quad ArabCulture (seed 42) & 90.9 & 90.9 & $\pm0.00$    & $[-3.03, +3.03]$ & 1.000 & 5 & 5  \\
\quad Palm (seed 0)         & 88.9 & 86.9 & $-2.02$      & $[-5.56, +1.52]$ & 0.388 & 4 & 8  \\
\quad Palm (seed 21)        & 89.9 & 84.3 & $-5.56^{*}$  & $[-9.60, -1.52]$ & \textbf{0.013} & 3 & 14 \\
\quad Palm (seed 42)        & 90.9 & 87.4 & $-3.54$      & $[-7.58, +0.51]$ & 0.143 & 5 & 12 \\
\quad FannOrFlop (seed 0)   & 88.9 & 88.4 & $-0.51$      & $[-3.54, +2.53]$ & 1.000 & 4 & 5  \\
\quad FannOrFlop (seed 21)  & 89.9 & 87.9 & $-2.02$      & $[-5.05, +0.51]$ & 0.289 & 2 & 6  \\
\quad FannOrFlop (seed 42)  & 90.9 & 90.4 & $-0.51$      & $[-4.04, +3.03]$ & 1.000 & 5 & 6  \\
\hline
\multicolumn{8}{l}{\textit{Fanar}} \\
\quad ArabCulture (seed 0)  & 87.9 & 88.4 & $+0.51$     & $[-2.02, +3.03]$ & 1.000 & 4 & 3 \\
\quad ArabCulture (seed 21) & 88.9 & 87.9 & $-1.01$     & $[-4.04, +2.02]$ & 0.754 & 4 & 6 \\
\quad ArabCulture (seed 42) & 90.4 & 87.9 & $-2.53$     & $[-6.06, +0.51]$ & 0.227 & 3 & 8 \\
\quad Palm (seed 0)         & 87.9 & 85.9 & $-2.02$     & $[-5.56, +1.52]$ & 0.388 & 4 & 8 \\
\quad Palm (seed 21)        & 88.9 & 85.4 & $-3.54$     & $[-7.07, -0.51]$ & 0.065 & 2 & 9 \\
\quad Palm (seed 42)        & 90.4 & 86.9 & $-3.54^{*}$ & $[-6.57, -1.01]$ & \textbf{0.039} & 1 & 8 \\
\quad FannOrFlop (seed 0)   & 87.9 & 89.4 & $+1.52$     & $[-2.02, +5.05]$ & 0.581 & 8 & 5 \\
\quad FannOrFlop (seed 21)  & 88.9 & 90.4 & $+1.52$     & $[-1.52, +5.05]$ & 0.549 & 7 & 4 \\
\quad FannOrFlop (seed 42)  & 90.4 & 89.4 & $-1.01$     & $[-4.55, +2.53]$ & 0.774 & 5 & 7 \\
\hline
\multicolumn{8}{l}{\textit{LLaMA}} \\
\quad ArabCulture (seed 0)  & 70.2 & 71.2 & $+1.01$ & $[-3.54, +5.56]$ & 0.824 & 11 & 9  \\
\quad ArabCulture (seed 21) & 67.2 & 68.7 & $+1.52$ & $[-3.03, +6.06]$ & 0.678 & 13 & 10 \\
\quad ArabCulture (seed 42) & 69.2 & 70.7 & $+1.52$ & $[-3.54, +6.57]$ & 0.690 & 14 & 11 \\
\quad Palm (seed 0)         & 70.2 & 74.7 & $+4.55$ & $[+0.00, +9.09]$ & 0.078 & 15 & 6  \\
\quad Palm (seed 21)        & 67.2 & 71.2 & $+4.04$ & $[-0.51, +9.09]$ & 0.152 & 16 & 8  \\
\quad Palm (seed 42)        & 69.2 & 70.2 & $+1.01$ & $[-3.54, +5.56]$ & 0.832 & 12 & 10 \\
\quad FannOrFlop (seed 0)   & 70.2 & 73.7 & $+3.54$ & $[-1.01, +8.08]$ & 0.210 & 15 & 8  \\
\quad FannOrFlop (seed 21)  & 67.2 & 69.7 & $+2.53$ & $[-2.53, +7.58]$ & 0.442 & 16 & 11 \\
\quad FannOrFlop (seed 42)  & 69.2 & 70.2 & $+1.01$ & $[-4.04, +6.06]$ & 0.845 & 14 & 12 \\
\hline
\multicolumn{8}{l}{\textit{Qwen}} \\
\quad ArabCulture (seed 0)  & 80.3 & 81.3 & $+1.01$   & $[-2.02, +4.04]$ & 0.727 & 5 & 3 \\
\quad ArabCulture (seed 21) & 80.3 & 79.8 & $-0.51$   & $[-3.03, +2.02]$ & 1.000 & 3 & 4 \\
\quad ArabCulture (seed 42) & 83.3 & 83.3 & $\pm0.00$ & $[-2.53, +2.53]$ & 1.000 & 3 & 3 \\
\quad Palm (seed 0)         & 80.3 & 82.3 & $+2.02$   & $[-0.51, +5.05]$ & 0.289 & 6 & 2 \\
\quad Palm (seed 21)        & 80.3 & 81.3 & $+1.01$   & $[-2.02, +4.04]$ & 0.754 & 6 & 4 \\
\quad Palm (seed 42)        & 83.3 & 85.4 & $+2.02$   & $[-0.51, +5.05]$ & 0.289 & 6 & 2 \\
\quad FannOrFlop (seed 0)   & 80.3 & 81.8 & $+1.52$   & $[-1.52, +5.05]$ & 0.549 & 7 & 4 \\
\quad FannOrFlop (seed 21)  & 80.3 & 83.3 & $+3.03$   & $[+0.00, +6.06]$ & 0.109 & 8 & 2 \\
\quad FannOrFlop (seed 42)  & 83.3 & 84.8 & $+1.52$   & $[-1.52, +4.55]$ & 0.508 & 6 & 3 \\
\hline
\end{tabular}
\caption{Improvement and regression breakdown on Jawaher proverbs across cultural fine-tuning (ArabCulture and Palm) and poetry fine-tuning (FannOrFlop). Base\% and FT\% are accuracy before and after fine-tuning; $\Delta\%$ is the percentage-point change, reported with 95\% paired-bootstrap confidence intervals over test items and exact McNemar $p$-values; Impr and Regr are the number of individual predictions improved or worsened. $^{*}$ marks $p<0.05$.}
\label{tab:per_file_proverbs_cultural_poetry}
\end{table*}

\begin{table}[t]
\centering
\small
\renewcommand{\arraystretch}{1.12}
\resizebox{\columnwidth}{!}{%
\begin{tabular}{lrcrr}
\hline
\textbf{Model / Dataset} & \textbf{$\Delta\%$} & \textbf{95\% CI} & \textbf{Impr} & \textbf{Regr} \\
\hline
\multicolumn{5}{l}{\textit{ALLaM}} \\
\quad Jawaher (seed 0)      & $+1.11$     & $[-3.33, +5.56]$ & 10 & 8  \\
\quad Jawaher (seed 21)     & $-1.11$     & $[-5.56, +3.33]$ & 8  & 10 \\
\quad Jawaher (seed 42)     & $-3.89$     & $[-7.78, +0.00]$ & 3  & 10 \\
\quad FannOrFlop (seed 0)   & $+1.11$     & $[-3.33, +5.56]$ & 9  & 7  \\
\quad FannOrFlop (seed 21)  & $-1.67$     & $[-5.56, +1.67]$ & 4  & 7  \\
\quad FannOrFlop (seed 42)  & $-5.00^{*}$ & $[-8.89, -1.11]$ & 2  & 11 \\
\hline
\multicolumn{5}{l}{\textit{Fanar}} \\
\quad Jawaher (seed 0)      & $+0.56$   & $[-3.89, +5.00]$ & 9 & 8 \\
\quad Jawaher (seed 21)     & $+0.56$   & $[-3.89, +5.00]$ & 8 & 7 \\
\quad Jawaher (seed 42)     & $\pm0.00$ & $[-3.33, +3.33]$ & 5 & 5 \\
\quad FannOrFlop (seed 0)   & $\pm0.00$ & $[-3.89, +3.89]$ & 7 & 7 \\
\quad FannOrFlop (seed 21)  & $\pm0.00$ & $[-4.44, +4.44]$ & 8 & 8 \\
\quad FannOrFlop (seed 42)  & $+2.22$   & $[-1.67, +6.11]$ & 8 & 4 \\
\hline
\multicolumn{5}{l}{\textit{LLaMA}} \\
\quad Jawaher (seed 0)      & $-0.56$   & $[-6.11, +5.00]$ & 12 & 13 \\
\quad Jawaher (seed 21)     & $-0.56$   & $[-5.56, +4.44]$ & 11 & 12 \\
\quad Jawaher (seed 42)     & $-1.11$   & $[-7.22, +5.00]$ & 15 & 17 \\
\quad FannOrFlop (seed 0)   & $+2.22$   & $[-3.33, +7.78]$ & 16 & 12 \\
\quad FannOrFlop (seed 21)  & $\pm0.00$ & $[-5.56, +5.56]$ & 13 & 13 \\
\quad FannOrFlop (seed 42)  & $-1.11$   & $[-7.22, +5.00]$ & 14 & 16 \\
\hline
\multicolumn{5}{l}{\textit{Qwen}} \\
\quad Jawaher (seed 0)      & $+1.67$ & $[-1.11, +4.44]$ & 5 & 2 \\
\quad Jawaher (seed 21)     & $+1.67$ & $[+0.00, +3.89]$ & 3 & 0 \\
\quad Jawaher (seed 42)     & $-0.56$ & $[-2.78, +1.67]$ & 2 & 3 \\
\quad FannOrFlop (seed 0)   & $+1.67$ & $[-1.67, +5.00]$ & 6 & 3 \\
\quad FannOrFlop (seed 21)  & $+1.11$ & $[-1.67, +3.89]$ & 4 & 2 \\
\quad FannOrFlop (seed 42)  & $-0.56$ & $[-3.33, +2.22]$ & 3 & 4 \\
\hline
\end{tabular}
}
\caption{Improvement and regression breakdown on AraDiCE-Culture across figurative fine-tuning (Jawaher) and poetry fine-tuning (FannOrFlop). $\Delta\%$ is the percentage-point change from base to fine-tuned accuracy, reported with 95\% paired-bootstrap confidence intervals over test items; Impr and Regr are the number of individual predictions improved or worsened. $^{*}$ marks $p<0.05$ under an exact McNemar test.}
\label{tab:per_file_aradice}
\end{table}

\begin{table}[h]
\centering
\small
\renewcommand{\arraystretch}{1.3}
\begin{tabular}{cp{4cm}}
\hline
\textbf{Count} & \textbf{Idiom} \\
\hline
\multicolumn{2}{c}{\textbf{Top 15 Most Frequently Improved}} \\
\hline
16 & \<دُودْ عَلَى عُودْ> \\
9  & \<عَلَى الْحَدِيدَهْ> \\
9  & \<أَكَلْ وِشهْ> \\
8  & \<عَلَى الْجِلْدَهْ> \\
6  & \<بِالْبَاعْ والدِّرَاعْ> \\
6  & \<إيدْ مِنْ وَرَا وِإيدْ مِنْ قُدَّامْ> \\
6  & \<كفه سايب> \\
6  & \<إيدُهْ نَاشْفَهْ> \\
6  & \<مِنْ هَبِّ وْمِنْ دَبّْ> \\
6  & \<وِشُّهْ يِقْطَعِ الْخَمِيرَهْ مِنِ البيتْ> \\
6  & \<بينُه وْبينُهْ مَا صَنَعِ الْحَدَّادْ> \\
6  & \<مَا يِنْزِلْشْ مِنِ الزُّورْ> \\
5  & \<طِلِعْ مِنْ عِينُهْ> \\
4  & \<الْخَالِقْ النَّاطِقْ> \\
4  & \<جَوَازِةْ نَصَارَى> \\
\hline
\multicolumn{2}{c}{\textbf{Top 15 Most Frequently Regressed}} \\
\hline
9 & \<زَيِّ النَّاسْ> \\
7 & \<خَلَّاهْ يِرِن> \\
6 & \<مُوشْ مِنْ تُوبُهْ> \\
6 & \<جَابْ دَاغُهْ> \\
6 & \<عَنْدُه الدُّنْيَا بِالْخُلْخَالْ> \\
6 & \<بِالْحِنْجِلْ وِالْمِنْجِلْ> \\
6 & \<جَابْهَا فِي قُبِّتُهْ> \\
6 & \<حَط في بَطْنُهْ بَطِّيخَهْ صِيفي> \\
6 & \<مُوشْ جَايِبْهَا الْبَر> \\
5 & \<الدُّنْيَا بِتِضْرَبْ وِتِقْلبْ> \\
5 & \<عَمَلِ البَحْرِ طْحِينَه> \\
5 & \<جَسِّ الْمَخَاضَهْ> \\
5 & \<إدَّارَى فِي ضِلِّ صُبَاعُهْ> \\
4 & \<تِلْتِ التَّلَاتَهْ> \\
4 & \<بَلَعْ رِيقُهْ> \\
\hline
\end{tabular}
\caption{Top 15 most frequently improved and regressed idioms across all models fine-tuned on cultural data (ArabCulture and Palm).}
\label{tab:top_improved_regressed_idioms}
\end{table}

\begin{table*}[h]
\centering
\small
\resizebox{0.7\textwidth}{!}{%
\begin{tabular}{cll}
\hline
\textbf{Count} & \textbf{Dialect} & \textbf{Proverb} \\
\hline
\multicolumn{3}{c}{\textbf{Top 15 Most Frequently Improved}} \\
\hline
10 & Mauritanian & \<الْمَاهُ وارِدْ امْعاكْ لا يْعَلَّكْلكْ.> \\
9  & Mauritanian & \<كل بخنوس افعين امو اغزال> \\
6  & Yemeni      & \<خزق وربك يرزق> \\
6  & Lebanese    & \<أكل الأخْضَر واليابس> \\
6  & Kuwaiti     & \<سو خير وقطه بحر.> \\
6  & Sudanese    & \<اب سن يضحك على اب سنين> \\
5  & Mauritanian & \<ساعت التغژاژ ما ينعراو الدروص> \\
4  & Algerian    & \<اللي ما رضى خببزة يرضى بنصها> \\
4  & Iraqi       & \<رجع أيد من وره وأيد من كدام> \\
4  & Jordanian   & \<إبْراسُهُ رِيشِه> \\
4  & Jordanian   & \<اللي ما يطول العنب حامض عنه يقول> \\
4  & MSA         & \<إذا أنت أكرمت الكريم ملكته وإن أنت أكرمت اللئيم تمردا> \\
4  & Yemeni      & \<العال في ثمنه> \\
4  & Egyptian    & \<إللي عَلى راسُه بَطحَة يِحَسِّس عَليها> \\
4  & Algerian    & \<الذي خرج من داره قل مقداره> \\
\hline
\multicolumn{3}{c}{\textbf{Top 15 Most Frequently Regressed}} \\
\hline
12 & Algerian    & \<الدار محلولة والمرأة مختولة> \\
8  & Kuwaiti     & \<الباب يوسع جمل> \\
8  & Lebanese    & \<حَدا بيشتري سمك ببحر؟> \\
7  & Egyptian    & \<عامل نفسه من بنها> \\
6  & Sudanese    & \<اسمع كلام الكبير و لو كان> \\
6  & Libyan      & \<البطن ما تجيب اعدوا.> \\
6  & Omani       & \<احابي حمد من اجل عيون محمد> \\
5  & Sudanese    & \<الكلب ينبح والجمل يفوت.> \\
4  & Sudanese    & \<القُفه ام اضنين بشيلوها نفرين.> \\
4  & Libyan      & \<اللى يُرقد يُرقد عليه زمانه> \\
4  & Saudi       & \<الدي مليح والسكات أحسن منو.> \\
4  & Jordanian   & \<شو جاب الزرقا للبلقا.> \\
4  & Egyptian    & \<أَبُو جُعْرَانْ فِي بِيتُهْ سُلْطَانْ> \\
4  & Moroccan    & \<آ المزوّق من برّا آش خبارك من دّاخل> \\
3  & Palestinian & \<اشتَرِ العَبد ولا ترَبّيه> \\
\hline
\end{tabular}
}
\caption{Top 15 most frequently improved and regressed proverbs across all fine-tuned models, with counts and dialect variety.}
\label{tab:top_improved_regressed_proverbs}
\end{table*}

\begin{table}[h]
\centering
\small
\resizebox{0.9\columnwidth}{!}{%
\begin{tabular}{lrrr}
\hline
\textbf{Dialect} & \textbf{Impr} & \textbf{Regr} & \textbf{Net} \\
\hline
Yemeni               & 10 &  2 & $+8$ \\
Mauritanian          & 11 &  3 & $+8$ \\
Jordanian            &  8 &  1 & $+7$ \\
Kuwaiti              &  6 &  0 & $+6$ \\
Tunisian             &  5 &  1 & $+4$ \\
MSA &  4 &  0 & $+4$ \\
Egyptian             &  7 &  4 & $+3$ \\
Bahraini                  &  2 &  0 & $+2$ \\
Moroccan             &  5 &  3 & $+2$ \\
Saudi                &  2 &  0 & $+2$ \\
Palestinian          &  7 &  6 & $+1$ \\
Syrian               &  5 &  4 & $+1$ \\
Iraqi                &  2 &  1 & $+1$ \\
Lebanese             &  5 &  4 & $+1$ \\
Libyan               &  2 &  2 & $0$ \\
Emirati              &  0 &  1 & $-1$ \\
Omani                  &  1 &  5 & $-4$ \\
Algerian             &  7 & 12 & $-5$ \\
Sudanese             &  5 & 13 & $-8$ \\
Qatari               &  3 & 11 & $-8$ \\
\hline
\end{tabular}
}
\caption{Net poetry fine-tuning effect by Arabic dialect variety on the proverb task.}
\label{tab:poetry_dialect}
\end{table}

\begin{table}[h]
\centering
\small
\resizebox{\columnwidth}{!}{%
\begin{tabular}{cp{5cm}}
\hline
\textbf{Count} & \textbf{Idiom} \\
\hline
\multicolumn{2}{c}{\textbf{Top 15 Most Frequently Improved}} \\
\hline
8 & \<دُودْ عَلَى عُودْ> \\
6 & \<مِنْ هَبِّ وْمِنْ دَبّْ> \\
4 & \<عَلَى الْجِلْدَهْ> \\
4 & \<الدُّنْيَا بِتِضْرَبْ وِتِقْلبْ> \\
3 & \<عَلَى الْحَدِيدَهْ> \\
3 & \<إيدْ مِنْ وَرَا وِإيدْ مِنْ قُدَّامْ> \\
3 & \<جَابْ دَاغُهْ> \\
3 & \<إتَّاوِب عَ النَّامُوسْ> \\
3 & \<بينُه وْبينُهْ مَا صَنَعِ الْحَدَّادْ> \\
3 & \<مَا يِنْزِلْشْ مِنِ الزُّورْ> \\
3 & \<وِشُّهْ يِقْطَعِ الْخَمِيرَهْ مِنِ البيتْ> \\
3 & \<أَكَلْ وِشهْ> \\
3 & \<إيدُهْ نَاشْفَهْ> \\
3 & \<طِلِعْ مِنْ عِينُهْ> \\
2 & \<حَطِّ صْبَاعُهْ فِي الشَّق> \\
\hline
\multicolumn{2}{c}{\textbf{Top 15 Most Frequently Regressed}} \\
\hline
5 & \<حَط في بَطْنُهْ بَطِّيخَهْ صِيفي> \\
4 & \<خَلَّاهْ يِرِن> \\
4 & \<عَمَلِ البَحْرِ طْحِينَه> \\
3 & \<مُوشْ مِنْ تُوبُهْ> \\
3 & \<أَكَلْ وِشهْ> \\
3 & \<زَيِّ النَّاسْ> \\
3 & \<جَابْهَا فِي قُبِّتُهْ> \\
3 & \<عَنْدُه الدُّنْيَا بِالْخُلْخَالْ> \\
2 & \<سَمَكْ لَبَنْ تَمْرْ هنْدِي> \\
2 & \<عَلَى الْجِلْدَهْ> \\
2 & \<لِعبْ بِالْبيضهْ وِالْحجَرْ> \\
2 & \<أَشْكرَهْ خَبَرْ> \\
2 & \<بِالرَّطْلْ> \\
2 & \<جَسِّ الْمَخَاضَهْ> \\
2 & \<كِلْمَه وْرَدّْ غَطَاها> \\
\hline
\end{tabular}
}
\caption{Top 15 most frequently improved and regressed idioms across all models fine-tuned on poetry data (FannOrFlop).}
\label{tab:top_improved_regressed_idioms_poetry}
\end{table}

\begin{table*}[h]
\centering
\small
\resizebox{0.7\textwidth}{!}{%
\begin{tabular}{cll}
\hline
\textbf{Count} & \textbf{Dialect} & \textbf{Proverb} \\
\hline
\multicolumn{3}{c}{\textbf{Top 15 Most Frequently Improved}} \\
\hline
4 & MSA          & \<إذا أنت أكرمت الكريم ملكته وإن أنت أكرمت اللئيم تمردا> \\
4 & Egyptian     & \<إللي عَلى راسُه بَطحَة يِحَسِّس عَليها> \\
3 & Mauritanian  & \<الْمَاهُ وارِدْ امْعاكْ لا يْعَلَّكْلكْ.> \\
3 & Algerian     & \<العدوة مزاح.> \\
3 & Yemeni       & \<خزق وربك يرزق> \\
3 & Syrian       & \<إذا في خير، ما كان تركه الطير.> \\
3 & Palestinian  & \<أَجَت الحَزِينة تِفرَح ما لاَقَتَش لهَا مِطرَح.> \\
3 & Moroccan     & \<اللهم العمش ولا العمى> \\
3 & Lebanese     & \<أكل الأخْضَر واليابس> \\
3 & Kuwaiti      & \<سو خير وقطه بحر.> \\
3 & Yemeni       & \<العرق دساس> \\
3 & Mauritanian  & \<كل بخنوس افعين امو اغزال> \\
3 & Sudanese     & \<اب سن يضحك على اب سنين> \\
2 & Jordanian    & \<إِبِنِ الْخَبَازِة ما يجوع> \\
2 & Algerian     & \<اللي ما رضى خببزة يرضى بنصها> \\
\hline
\multicolumn{3}{c}{\textbf{Top 15 Most Frequently Regressed}} \\
\hline
6 & Qatari       & \<التجدي ولا العمى> \\
6 & Algerian     & \<الدار محلولة والمرأة مختولة> \\
3 & Palestinian  & \<اذا لَم تَكُن فَارِسًا كُنتَ الفَرِيسَة> \\
3 & Sudanese     & \<الطشاش في بلد العمي شوف> \\
3 & Egyptian     & \<عامل نفسه من بنها> \\
3 & Qatari       & \<أمن ولا اتخون> \\
3 & Mauritanian  & \<كل ديقة وراها صحبة> \\
3 & Syrian       & \<أنا ويّاك والزمن طويل> \\
3 & Omani        & \<احابي حمد من اجل عيون محمد> \\
2 & Sudanese     & \<اسمع كلام الكبير و لو كان> \\
2 & Algerian     & \<العدوة مزاح.> \\
2 & Libyan       & \<البطن ما تجيب اعدوا.> \\
2 & Sudanese     & \<الكلب ينبح والجمل يفوت.> \\
2 & Lebanese     & \<أكلا عَ بارد المستريح> \\
2 & Moroccan     & \<آ المزوّق من برّا آش خبارك من دّاخل> \\
\hline
\end{tabular}
}
\caption{Top 15 most frequently improved and regressed proverbs across all models fine-tuned on poetry data (FannOrFlop), with counts and dialect variety.}
\label{tab:top_improved_regressed_proverbs_poetry}
\end{table*}

\section{Use of AI Assistants}
\label{app:ai-assistants}

LLM-based writing assistants were used to support writing and editing of the manuscript, including suggestions for rephrasing and refinement. All research design, analysis, interpretation, and final wording decisions were made by the authors.

\end{document}